\documentclass{article}

\PassOptionsToPackage{numbers, compress}{natbib}
\usepackage[preprint]{neurips_2025}

\usepackage{authblk} 
\usepackage[utf8]{inputenc} 
\usepackage[T1]{fontenc}    
\usepackage{hyperref}       
\usepackage{url}            
\usepackage{booktabs}       
\usepackage{amsfonts}       
\usepackage{nicefrac}       
\usepackage{microtype}      
\usepackage{xcolor}         
\usepackage{graphicx}  
\usepackage{float}  
\usepackage{subfigure}  
\usepackage{caption} 
\usepackage{adjustbox}
\usepackage{amsmath,amssymb}
\usepackage{wrapfig}
\usepackage{lipsum} 
\usepackage{booktabs}
\usepackage{amsfonts}       
\usepackage{siunitx}        
\usepackage{enumitem}
\usepackage{calc}
\usepackage{textcomp}
\usepackage{multirow}
\usepackage{afterpage}     
\usepackage{booktabs}
\makeatletter
\def\@fnsymbol#1{\ensuremath{\ifcase#1\or *\or \dagger\or \ddagger\or \S\or \P\else\@ctrerr\fi}}
\makeatother

\title{ChronoTailor: Harnessing Attention Guidance for Fine-Grained Video Virtual Try-On}
\author{
  Jinjuan Wang$^{1,\ast}$,
  Wenzhang Sun$^{2,\ast}$,
  Ming Li$^{1}$,
  Yun Zheng$^{1}$,
  Fanyao Li$^{1}$,
  Zhulin Tao$^{1,\dagger}$,
  Donglin Di$^{2}$,
  Hao Li$^{2}$,
  Wei Chen$^{2}$,
  Xianglin Huang$^{1}$\\
  $^{1}$Communication University of China,
  $^{2}$Li Auto \\
}

\begin{document}
\captionsetup[figure]{skip=10pt} 
\maketitle
\footnotetext{$^{\ast}$Equal contribution}
\footnotetext{$^{\dagger}$Corresponding author: \texttt{taozhulin@gmail.com}}
\begin{abstract}\label{sec:abs}

Video virtual try-on aims to seamlessly replace the clothing of a person in a source video with a target garment. Despite significant progress in this field, existing approaches still struggle to maintain continuity and reproduce garment details. In this paper, we introduce ChronoTailor, a diffusion-based framework that generates temporally consistent videos while preserving fine-grained garment details. By employing a precise spatio-temporal attention mechanism to guide the integration of fine-grained garment features, ChronoTailor achieves robust try-on performance.  
First, ChronoTailor leverages region-aware spatial guidance to steer the evolution of spatial attention and employs an attention-driven temporal feature fusion mechanism to generate more continuous temporal features. This dual approach not only enables fine-grained local editing but also effectively mitigates artifacts arising from video dynamics.
Second, ChronoTailor integrates multi-scale garment features to preserve low-level visual details and incorporates a garment-pose feature alignment to ensure temporal continuity during dynamic motion. Additionally, we collect StyleDress, a new dataset featuring intricate garments, varied environments, and diverse poses, offering advantages over existing public datasets, and will be publicly available for research. Extensive experiments show that ChronoTailor maintains spatio-temporal continuity and preserves garment details during motion, significantly outperforming previous methods.


\end{abstract}
\section{Introduction}\label{sec:intro}
Driven by e-commerce expansion and demands for immersive digital experiences, video virtual try-on technology \cite{nguyen2024swifttry,fang2024vivid} enables seamless garment replacement in dynamic videos, allowing users to visualize wearing target clothing without physical trials. By enhancing immersion and interactivity, it bridges the gap between static product images and real-world wearing scenarios. A key challenge is preserving fine details like textures, patterns, and fabric folds while ensuring consistent visuals across frames—especially when the body moves in complex poses.

Previous methods use flow-driven \cite{wang2018toward,Dong_Liang_Shen_Wu_Chen_Yin_2019,xu2024tunnel} warping modules and neural generators to align garments via shape deformation. However, their reliance on warping operations causes inherent limitations in temporal coherence.
Recent approaches leverage pre-trained diffusion models \cite{zhu2023tryondiffusion,chen2024anydoor,morelli2023ladi,gou2023taming,Kim_Gu_Park_Park_Choo_2023}, such as dual-UNet architectures \cite{ronneberger2015u} and Diffusion Transformer-based frameworks \cite{jiang2024fitdit,zeng2024cat}, to eliminate explicit warping and integrate temporal attention layers for modeling motion dynamics. While these methods enhance generative scalability and partially address coherence issues, key challenges persist: (1) Spatial-temporal inconsistency of injected garment features during dynamic motion, leading to misalignment or flickering artifacts; (2) Degradation of fine-grained details due to over-smoothing in generative processes and insufficient high-fidelity training data.
We claim that virtual video try-on requires accurate guidance to steer the injection of garment information, and such information should stably preserve fine-grained garment details during motion.
\begin{figure}
    \centering
    \includegraphics[width=\textwidth]{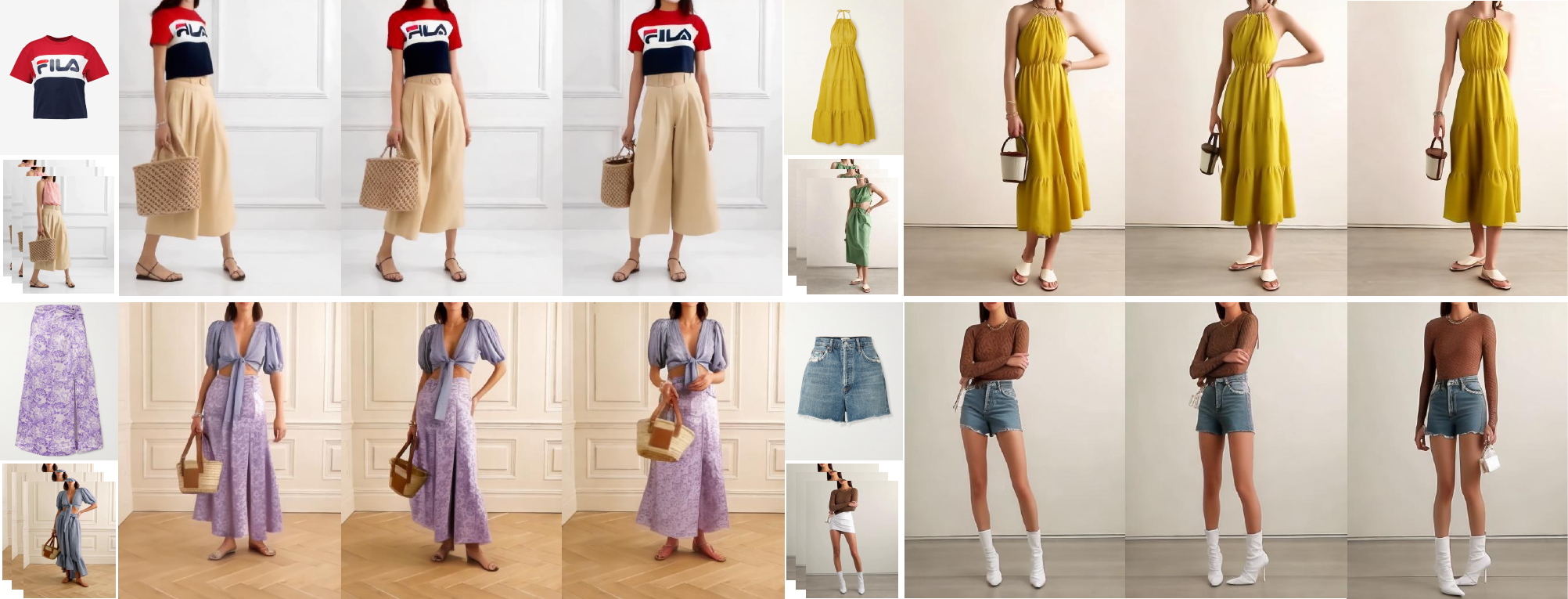}
    \caption{ChronoTailor excels at generating high-quality virtual try-ons for diverse garments, effectively maintaining visual coherence and preserving background details.}
    \label{First Image}
    \vspace{-10mm}
\end{figure}

In this paper, we present ChronoTailor, a diffusion-based \cite{ho2020denoising}framework designed to provide stable spatial-temporal attention guidance and pose-aligned garment features with fine-grained details for precise information injection. ChronoTailor uses segmentation results to guide attention to target regions and incorporates an additional attention regularization term to balance spatial consistency between edited and unedited areas. To enhance temporal stability, we deploy asymmetric cross-attention to model temporal dependencies in video sequences. Moreover, the framework extracts multi-level garment features to preserve intricate details and aligns these features with pose representations via cross-attention modules with shared weights, ensuring coherent interaction between clothing and human dynamics.

Specifically, ChronoTailor is composed of two key components: (1) \textit{Spatial-Temporal Attention Guidance}, which provides a stable spatial-temporal guidance for garment injection. First, we employ segmentation masks \cite{xiong2024sam2} to guide the evolution of the spatial attention map. With the aim of attaining an equilibrium of information within and outside the edited region, we further incorporate a regularization term specifically for the attention map. This incorporation empowers the information injection process to place additional emphasis on unedited regions. In addition, we employ an asymmetric cross-space attention mechanism, using original features as
queries and features obtained by randomly fusing the current frame with other frames within the video clip as key-values, to effectively integrate temporal information. (2) \textit{Multi-scale Garment-Pose Feature Alignment}, which enables the acquisition of more precise garment information during human movement. We extract multi-level garment features and fuse them with learnable weights to enhance low-level garment details. Additionally, to align garments with poses, the cross-attention modules in the reference and denoising UNets integrate pose and garment features respectively and align them via weight sharing. Additionally, we collect \textit{StyleDress}, a new dataset featuring intricate garments, varied environments, and diverse human poses, comprising 14,258 high-quality images and 12,500 videos, offering advantages over existing public datasets. StyleDress will be publicly available for research. In conclusion, the main contributions can be summarized as follows:
\begin{itemize}[labelsep=0.4em, leftmargin=1em, itemindent=0em]
    \vspace{-8pt}
    \setlength{\itemsep}{0pt}
    \setlength{\parsep}{0pt}
    \setlength{\parskip}{0pt}
    \item We introduce ChronoTailor, a diffusion-based framework that leverages spatial-temporal attention guidance and fine-grained pose-aligned garment features to effectively generate temporally coherent videos while preserving intricate garment details, as shown in Figure \ref{First Image}.
    \item We introduce Spatial-Temporal Attention Guidance and Multi-scale Garment-Pose Feature Alignment, enabling the model to focus more on edited regions and achieve better preservation of garment textures during motion.
    \item We present StyleDress, a comprehensive dataset featuring intricate garments, diverse environments, and varied human poses. Outperforming existing datasets in diversity and quality, it will be publicly released to advance research.
\end{itemize}

\section{Related Work}\label{sec:related work}
\vspace{-1mm}
\subsection{Image Virtual Try-On}
Traditional virtual try-on methods, predominantly Generative Adversarial Networks \cite{goodfellow2014generative,Dong_Liang_Shen_Wu_Chen_Yin_2019,Han_Wu_Wu_Yu_Davis_2018,Choi_Park_Lee_Choo_2021,goodfellow2014generative} (GANs), synthesize images of individuals wearing desired clothing by deforming garments and compositing them onto a person's image while aiming to preserve identity and consistency. Early works like VITON \cite{Han_Wu_Wu_Yu_Davis_2018} used coarse-to-fine strategies for garment deformation, while CP-VTON \cite{Ge_Song_Zhang_Ge_Liu_Luo_2021} improved alignment via geometric matching. VITON-HD \cite{Choi_Park_Lee_Choo_2021} introduced the ALIAS mechanism to mitigate misalignment issues. However, these GAN-based approaches struggle with generalization to complex poses and diverse backgrounds \cite{honda2019viton}. 
The emergence of diffusion models has revolutionized the artificial intelligence generated content \cite{sun2024uniavatar,zhang2025semanticlatentmotionportrait,liu2025moee}field and inspired innovative virtual try-on techniques \cite{Choi_Kwak_Lee_Choi_Shin_2024,esser2023structure,Kim_Gu_Park_Park_Choo_2023,zhu2023tryondiffusion,li2022diffcloth,baldrati2023multimodal}. StableVITON \cite{Kim_Gu_Park_Park_Choo_2023} employs zero-cross attention to improve texture fidelity but lacks explicit multi-scale modeling of garment features, resulting in suboptimal adaptive fusion of details across different resolutions. The parallel UNet architecture proposed in TryOnDiffusion \cite{zhu2023tryondiffusion} captures multi-scale information in an implicit manner via shared parameters. This design lacks the capability to dynamically adjust feature weights in response to the complexity of input garments. ChronoTailor pioneers a learnable cross-scale weighting mechanism to capture both local textures and global semantics simultaneously, enabling adaptive feature fusion and addressing the detail loss inherent in traditional static feature aggregation.
\subsection{Video Virtual Try-On}
\vspace{-1mm}
Current video virtual try-on methods exhibit relatively primitive applications of attention mechanisms\cite{cui2021dressing,chen2021fashionmirror,kuppa2021shineon,wang2018toward,Jiang_Wang_Yan_Liu_Bigo,wei20253dv,chen2024livephoto,hu2022make}. For example, the visual Transformer in ClothFormer \cite{Jiang_Wang_Yan_Liu_Bigo} models garment-human relationships within single frames exclusively via self-attention, lacking explicit semantic mask guidance. This leads to attention weights diffusing into non-editing regions, causing boundary artifacts. Although Tunnel TryOn \cite{Xu_Zhang_Liew_Yan_Liu_Zhang_Feng_Shou} pioneered the use of diffusion models for video try-on, the generated boundaries between garment and non-garment regions remain blurred, with noticeable feature leakage. ChronoTailor mitigates such artifacts by balancing contextual information across regions and introduces an attention-driven temporal feature fusion mechanism to effectively alleviate error accumulation in long videos.

Additionally, existing methods inadequately modeled the correlation between reference dynamics and human poses\cite{karras2023dreampose,hu2024animate,zhu2024champ,wei20253dv}. FashionMirror \cite{chen2021fashionmirror} and MV-TON \cite{Zhong_Wu_Tan_Lin_Wu_2021} propagated pose-agnostic global features solely through optical flow or memory modules, failing to explicitly disentangle garment deformation from joint movements. This resulted in garment tearing during dynamic actions (e.g., sleeve texture fractures when swinging arms). While ViViD \cite{fang2024vivid} introduced pose information as conditional inputs, it fused them via simplistic fusion mechanisms, unable to establish spatial correspondences between garment features and pose coordinates.
ChronoTailor achieves cross-level alignment for the first time by integrating global-local pose information with fine-grained garment features through a dual cross-attention mechanism, guiding accurate deformation of garment features.
\vspace{-2mm}
\section{Method}\label{sec:method}
\vspace{-2mm}
\begin{figure*}
    \centering
    \includegraphics[width=\textwidth]{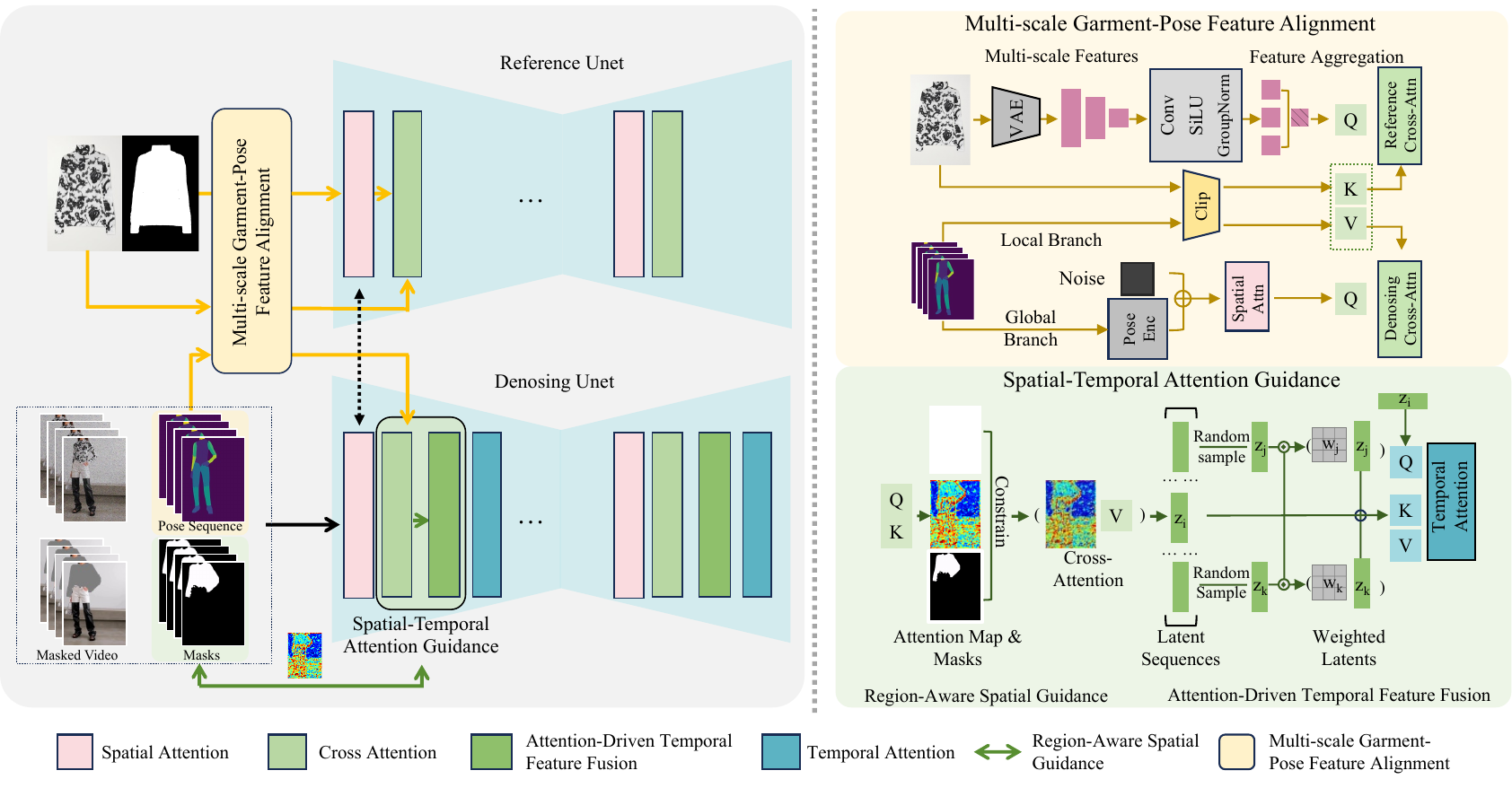}
    \caption{\textbf{Architecture of the ChronoTailor.} Spatial-Temporal Attention Guidance enables the acquisition of stable guidance for garment feature injection. Multi-scale Garment-Pose Feature Alignment, meanwhile, facilitates the capture of more precise garment information during motion.}
    \vspace{-4 mm}
    \label{arch}
\end{figure*}
\paragraph{Overall Architecture.}
The architecture, illustrated in Figure \ref{arch}, consists of two main components: (1) Spatial-Temporal Attention Guidance, which includes Region-Aware Spatial Guidance (RASG) that uses segmentation masks to guide spatial attention map evolution, and Attention-Driven Temporal Feature Fusion (ATFF), which employs an asymmetric cross-space attention mechanism using original features as queries and randomly fused other frame features as key-values to integrate temporal information effectively. (2) Multi-scale Garment-Pose Feature Alignment, which first applies Adaptive Multi-scale Feature Extraction (AMFE) to enhance low-level garment details, and then aligns garment features with pose features using Garment-Pose Feature Alignment (GPFA). More details are available in \ref{supp:Preliminary}.
\vspace{-1mm}
\subsection{Spatial-Temporal Attention Guidance}
\vspace{-1mm}
In the process of feature integration, the guidance of attention is critical. The goal of spatial-temporal attention guidance is to adjust the regions of attention from both spatial and temporal dimensions, providing more accurate guidance for the feature integration process.
\subsubsection{Region-Aware Spatial Guidance}
Conventional cross-attention mechanisms typically lack explicit semantic guidance, resulting in diffuse attention across replacement regions. To address these limitations, we introduce Region-Aware Spatial Guidance (RASG), which leverages segmentation masks to explicitly guide attention toward edit regions. Furthermore, we incorporate a regularization term to balance attention distribution, preventing overemphasis on edited regions and ensuring that critical contextual details beyond the replacement area are preserved for spatial consistency. RASG is formulated as:
\begin{equation}
\mathcal{L}_{\text{RASG}} = \sum_{i \in \mathcal{N}} \sum_{a \in \mathcal{A}} \| \mathcal{M} - A_i^{(a)} \|_2^2 + \lambda_{\mathcal{N}} \sum_{i \in \mathcal{N}} \sum_{a \in \bar{\mathcal{A}}} \| \mathbf{1} - A_i^{(a)} \|_2^2 ,
\end{equation}
where $\mathcal{N}$ denotes the video sequence length, $\mathcal{M}$ represents the segmentation mask, $A_i^{(a)}$ is the attention probability for location $a$ in frame $i$, and $\lambda_{\mathcal{N}}$ balances the contributions of positive and negative attention regularization. The first term enforces alignment between attention maps and the target region, while the second term regularizes attention in non-target regions to maintain global context. The final loss function integrates RASG with the base diffusion model:
\begin{equation}
\mathcal{L}_{\text{ChronoTailor}} = \mathcal{L}_{\text{LDM}} + \lambda_{\text{R}} \mathcal{L}_{\text{RASG}} ,
\end{equation}
where $\mathcal{L}_{\text{LDM}}$ is the loss of the latent diffusion model and $\lambda_{\text{R}}$ tunes the weight of RASG. This dual-objective strategy enables the model to focus precisely on target garment regions using segmentation guidance; preserve global coherence by leveraging contextual information from unedited areas.

\subsubsection{Attention-Driven Temporal Feature Fusion}
Attention-Driven Temporal Feature Fusion (ATFF) employs an asymmetric cross-space attention mechanism, using original features as queries and features obtained by randomly fusing the current frame with other frames within the video clip as key-values, to effectively integrate temporal information. Given garment features $\{z_{1}, z_{2}, \ldots, z_{N}\}$ from a video clip, for each current frame feature $z_i$, we compute attention scores $w_{j,k}$ by randomly selecting frames from the entire clip instead of strictly adjacent frames. Specifically:
\begin{equation}
w_{j,k} = \mathit{Softmax}\left( \sum_{d=-1} (z_i \cdot z_{j,k}) \right) 
\end{equation}
where $d=-1$ denotes the dimensional index for feature aggregation along the last dimension. These scores weight the randomly selected features $z_j$ and $z_k$ for fusion with $z_i$, yielding:
\begin{equation}
    \overline{z}=\mathit{GroupNorm}(w_{j}\cdot z_{j}+z_i+w_{k}\cdot z_{k}) .
\end{equation}
We employ a cross-attention layer where $\overline{z}$ is used as keys/values and $z_i$ as queries. Through linear projections with shared weights, the corresponding keys $\overline{K}$, values $\overline{V}$, and queries $Q$ are computed.
The final output of ATFF is calculated as:
\begin{equation}
    \hat{z}_i=\mathit{Softmax}\left(\frac{Q\overline{K}^T}{\sqrt{d_k}}\right)\overline{V},
\end{equation}
where $d_k$ is the dimensionality of the key vectors. This design thereby enhances temporal continuity.
\subsection{Multi-scale Garment-Pose Feature Alignment}
Precise garment detail preservation and accurate pose alignment are critical challenges in virtual try-on. We propose Adaptive Multi-scale Feature Extraction (AMFE) for optimized garment features and Garment-Pose Feature Alignment (GPFA) for precise alignment, enhancing virtual try-on quality.
\subsection{Adaptive Multi-scale Feature Extraction}
Adaptive Multi-scale Feature Extraction (AMFE) addresses a critical limitation of conventional methods—their exclusive reliance on single-level feature representations, which fails to capture garment details. AMFE extracts hierarchical garment features by processing feature maps ${f}_l$ from the VAE's last 3 downsampling layers using a structured adaptive fusion. Initially, a standardized transformation $\mathcal{T}$ is applied to each feature map ${f}_l$:
\begin{equation}
    \bar{f}_{l} = \mathcal{T}({f}_{l}) = \mathcal{S} \circ \mathcal{N} \circ \mathcal{W}_{2d} \circ \mathcal{P}({f}_{l}) ,
\end{equation}
where $l \in \{1, 2, 3\}$ indexes the VAE downsampling layers, $\mathcal{P}$ aligns resolution via pixel-shuffling, $\mathcal{W}_{2d}$ mixes channels with $1 \times 1$ convolution, $\mathcal{N}$ performs group normalization, and $\mathcal{S}$ is the SiLU activation. This cascade normalizes features for consistent fusion. Subsequently, AMFE employs a dynamic fusion strategy with learnable weights $\vec{\alpha}=[\alpha_{1}, \alpha_{2}, \alpha_{3}]$:
\begin{equation}
\hat{f} = \alpha_{1} \cdot \mathcal{F}_1(\bar{f}_{1}) + \alpha_{2} \cdot \mathcal{F}_2(\bar{f}_{2}) + \alpha_{3} \cdot \bar{{f}}_{3} ,
\end{equation}
where $\mathcal{F}_1$ uses $3 \times 3$ convolutions to capture local details, $\mathcal{F}_2$ employs $7 \times 7$ convolutions for global semantics, and $\bar{{f}}_{3}$ represents the original output of the VAE encoder. The weights $\vec{\alpha}$ are optimized during training to dynamically balance contributions from different scales, ensuring retention of critical details.

\subsection{Garment-Pose Feature Alignment}
Leveraging the texture representations from AMFE, we introduce the Garment-Pose Feature Alignment (GPFA) module to address the challenge of aligning garments with human poses. GPFA integrates local and global pose information to establish precise correspondences between garment features and body configurations, ensuring semantic consistency across dynamic movements.

First, we extract dense pose sequences \(x_p\) using DensePose and derive global pose features \(f_p\) via a pre-trained CLIP image encoder. For local pose modeling, \(x_p\) is processed through a four-layer learnable convolutional network \(\text{Conv}_\theta\), then fused with a noise latent feature \(z_0\) to form the input \(z\) for the denoising U-Net:
\begin{equation}
z = z_0 + \text{Conv}_\theta(x_p),
\end{equation}
where \(\text{Conv}_\theta\) denotes a convolutional layer with learnable parameters \(\theta\). To establish cross-modal dependencies, we design cross-attention modules in both reference and denoising UNet architectures with a weight-sharing mechanism \(W\). The reference cross-attention's query \(Q_R\), key \(K_R\), and value \(V_R\) are defined as:
\begin{equation}
Q_R = z_R W_Q, \quad K_R = f_g W_K, \quad V_R = f_g W_V,
\end{equation}
where \(W_Q, W_K, W_V\) are shared attention weight matrices; \(f_g\) is the CLIP-encoded garment feature; and \(z_R\) is the latent feature from the reference UNet. Similarly, the denoising cross-attention operates as:
\begin{equation}
Q_D = z_D W_Q, \quad K_D = f_p W_K, \quad V_D = f_p W_V,
\end{equation}
with \(f_p\) denoting the CLIP-encoded pose feature and \(z_D\) the latent feature from the denoising UNet. By sharing weights, GPFA enhances cross-modal generalization, thus maintaining garment stability during dynamic motion.
\section{StyleDress}\label{sec:styledress}
\begin{figure}[t]
\centering\scriptsize
\begin{minipage}[h]{0.48\textwidth} 
\centering\scriptsize
    \includegraphics[width=\linewidth]{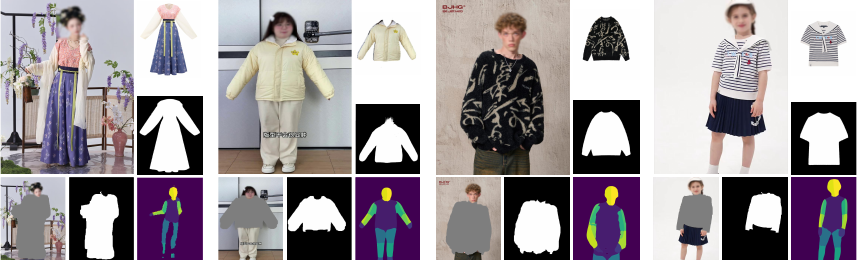}
    \caption{\normalsize Sample pairs from the StyleDress.}
    \label{fig:dataset_examples}
\end{minipage}%
\hfill 
\begin{minipage}[h]{0.48\textwidth}
\centering
\captionsetup{type=table, font=normalsize}
\caption{\normalsize Comparison of StyleDress with public datasets.}
\label{tab:styledress_comparison}
\setlength{\tabcolsep}{1.8pt}
\begin{tabular}{>{\scriptsize\centering\arraybackslash}c>{\scriptsize\centering\arraybackslash}c>{\scriptsize\centering\arraybackslash}c>{\scriptsize\centering\arraybackslash}c>{\scriptsize\centering\arraybackslash}c>{\scriptsize\centering\arraybackslash}c>{\scriptsize\centering\arraybackslash}c}
        \hline
        \renewcommand{\arraystretch}{3}
        \raisebox{0.7ex}{\strut}Dataset\raisebox{0.7ex}{\strut} & 
        \raisebox{0.7ex}{\strut}Pairs\raisebox{0.7ex}{\strut} & 
        \raisebox{0.7ex}{\strut}Media\raisebox{0.7ex}{\strut} & 
        \raisebox{0.7ex}{\strut}Gender\raisebox{0.7ex}{\strut} & 
        \raisebox{0.7ex}{\strut}Body\raisebox{0.7ex}{\strut} & 
        \raisebox{0.7ex}{\strut}Pose\raisebox{0.7ex}{\strut} & 
        \raisebox{0.7ex}{\strut}Background\raisebox{0.7ex}{\strut} \\
        \hline
        \renewcommand{\arraystretch}{2}
        
        VVT & 791 & V & F & Avg. & Catwalk & Indoor \\
        ViViD & 9700 & V & F & Avg. & Catwalk & Indoor \\
        \textbf{StyleDress} & 26758 & I-V & F-M & Diverse & Diverse & Diverse \\
        \hline
    \end{tabular}
\end{minipage}
\end{figure}

Public image/video virtual try-on datasets face challenges: VITON-HD \cite{lee2022high} and DressCode \cite{morelli2022dress} offer high-quality visuals but suffer from simple backgrounds and limited poses. VVT \cite{Dong_Liang_Shen_Wu_Chen_Yin_2019}(video-based) has monotonous actions, uniform backgrounds, and low resolution (\(256 \times 192\)), restricting real-world use. ViViD \cite{fang2024vivid}provides higher resolution (\(832 \times 624\)) and more clothing items but exhibits gender imbalance, singular poses, and limited body shape diversity (predominantly slender females). Real-world scenarios demand diverse backgrounds, camera angles, and lighting conditions, necessitating robust datasets. To bridge this gap, we present StyleDress, a high-resolution video dataset for virtual try-on featuring: complex backgrounds, varied body shapes/poses, and balanced gender representation (Figure \ref{fig:dataset_examples}) and Enhanced practical applicability. Table \ref{tab:styledress_comparison} compares dataset diversity, with additional visualizations in Appendix \ref{supp:StyleDress}.

\section{Experiments}\label{sec:exp}
\paragraph{Datasets.}
\begin{figure}[t] 
    \centering
    \includegraphics[width=\textwidth]{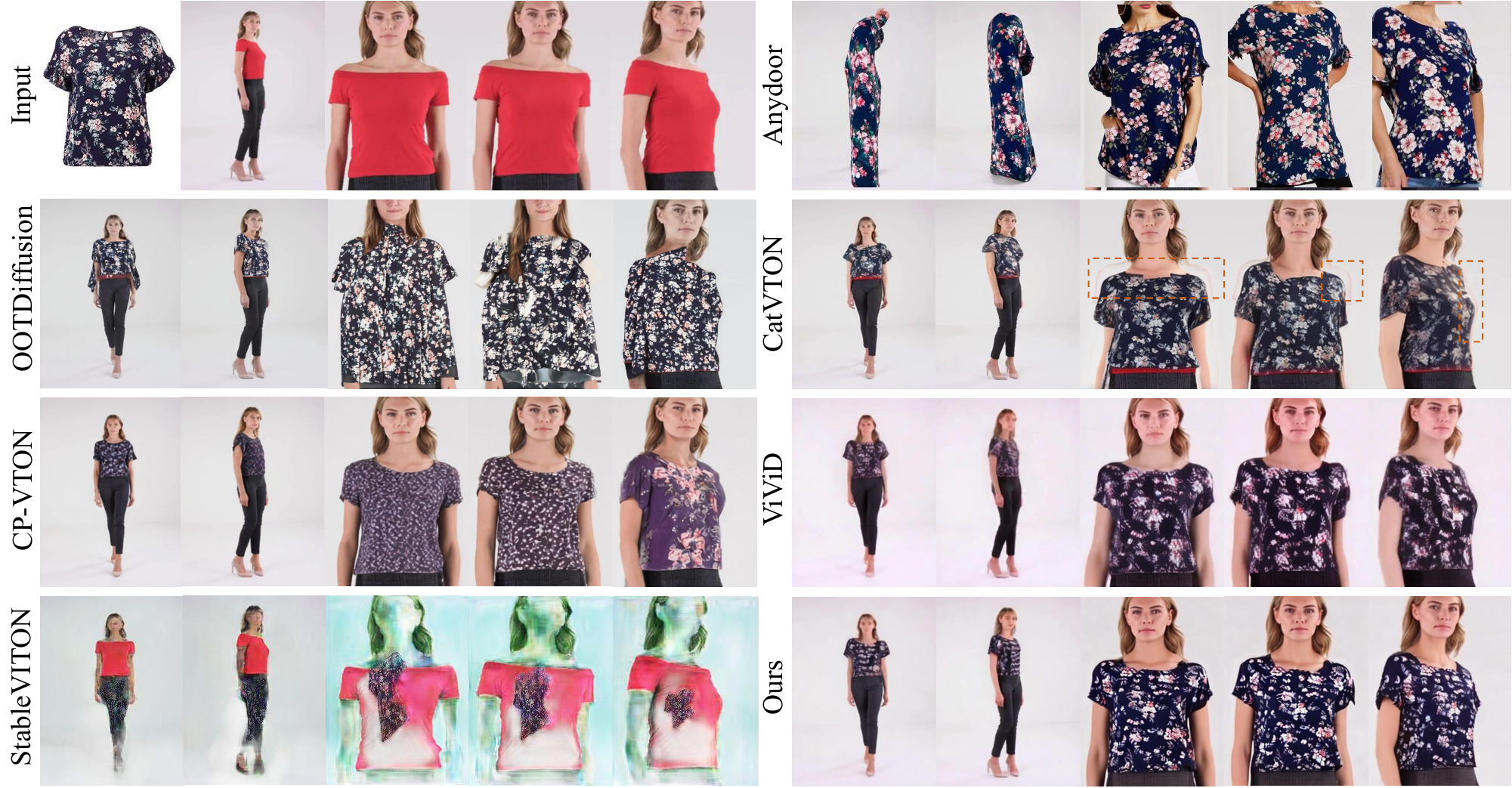}
    \caption{Qualitative comparison on the VVT dataset. The dashed box highlights an area with barely perceptible artifacts. Our method demonstrates superior temporal consistency (reduced artifacts) and generates finer garment details.}
    \vspace{-5mm}
    \label{vvt}
\end{figure}
\begin{figure}[t] 
    \centering
    \includegraphics[width=\textwidth]{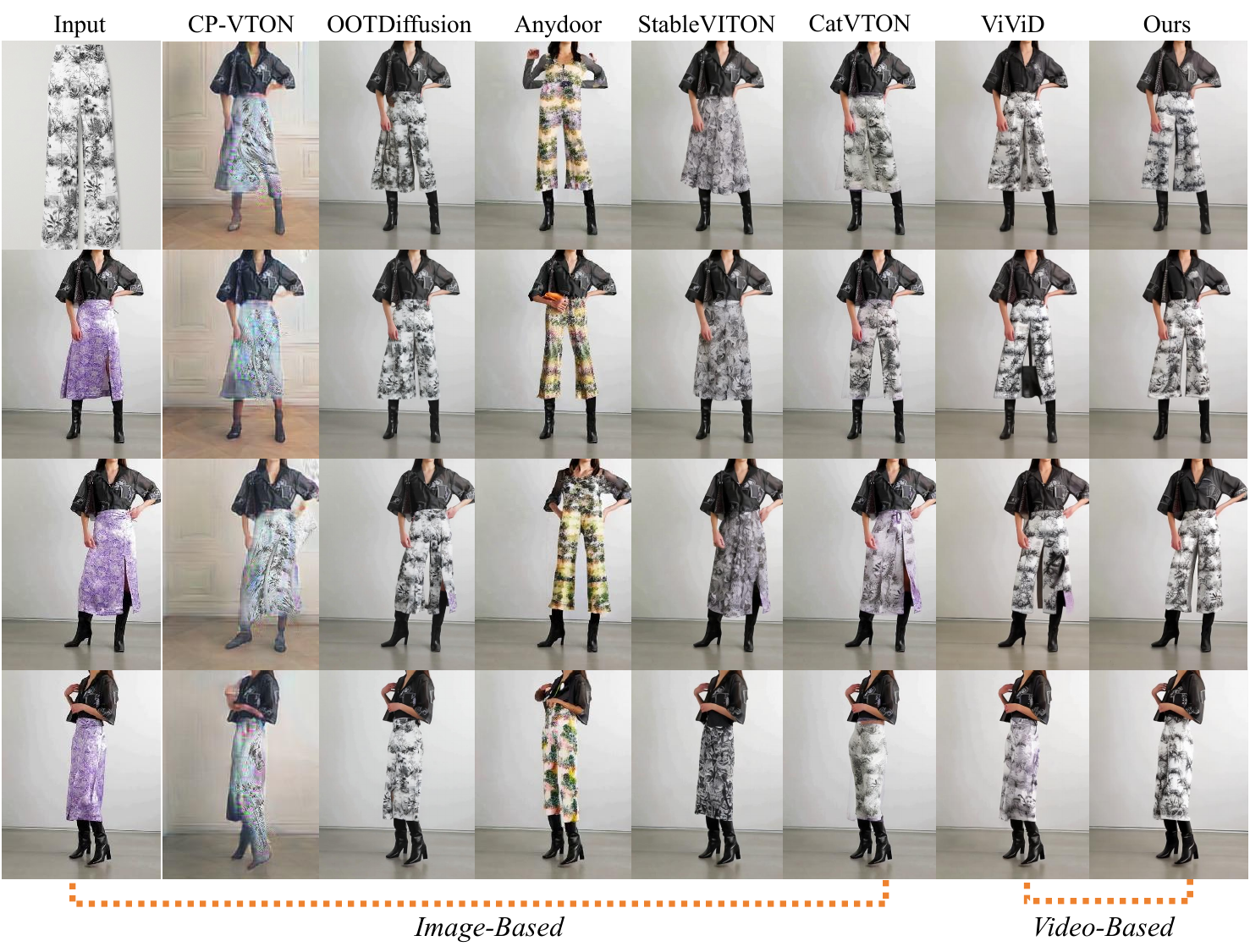}
    \caption{Qualitative comparison on the ViViD dataset. We demonstrate the results for lower garments, where ChronoTailor better preserves the patterns and shapes of trousers.}
    \label{visual_vivid}
    \vspace{-4mm}
\end{figure}
\begin{table}[t]
\begin{minipage}[c]{0.48\textwidth}
\centering
    \captionsetup{type=table}
    \caption{Comparison on the VVT dataset: $\uparrow$ denotes higher is better, while $\downarrow$ indicates lower is better.}
    \resizebox{\columnwidth}{!}{
     \large
         \begin{tabular}{lccccccccccc}
         \toprule
        \renewcommand{\arraystretch}{1.8} 
         Method& SSIM$\uparrow$ & LPIPS$\downarrow$ & $VFID_{I3D}$$\downarrow$ & $VFID_{ResNeXt}$$\downarrow$ \\
            \midrule
                 CP-VTON \cite{wang2018toward} & 0.459 & 0.535 & 6.361 & 12.100 \\
                FW-GAN \cite{Dong_Liang_Shen_Wu_Chen_Yin_2019} & 0.675 & 0.283 & 8.019 & 12.150 \\
                 PBAFN \cite{Ge_Song_Zhang_Ge_Liu_Luo_2021} & 0.870 & 0.157 & 4.516 & 8.690 \\
                MVTON \cite{Zhong_Wu_Tan_Lin_Wu_2021} & 0.853 & 0.068 & 8.367 & 9.702 \\
                 Tunnel Try-on \cite{xu2024tunnel} & 0.913 & 0.054 & \textbf{3.345} & 4.614 \\
                ClothFormer \cite{Jiang_Wang_Yan_Liu_Bigo} & 0.921 & 0.081 & 3.967 & 5.048 \\
                AnyDoor \cite{chen2024anydoor} & 0.800 & 0.127 & 4.535 & 5.990 \\
                StableVITON \cite{Kim_Gu_Park_Park_Choo_2023} & 0.760 & 0.184 & 17.068 & 11.254 \\
                CatVTON \cite{chong2024catvton} & 0.887 & 0.099 & 8.311 & 6.013 \\
                 ViViD \cite{fang2024vivid} & 0.949 & 0.068 & 3.405 & 5.074 \\
                \textbf{Ours} & \textbf{0.978} & \textbf{0.049} & 3.721 & \textbf{4.065} \\
            \bottomrule
        \label{VVT_DATA}
        \end{tabular}
        }
\label{AMFE}
\end{minipage}
\hfill
\begin{minipage}[c]{0.47\textwidth}
\centering
    \captionsetup{type=figure}
    \includegraphics[width=\textwidth]{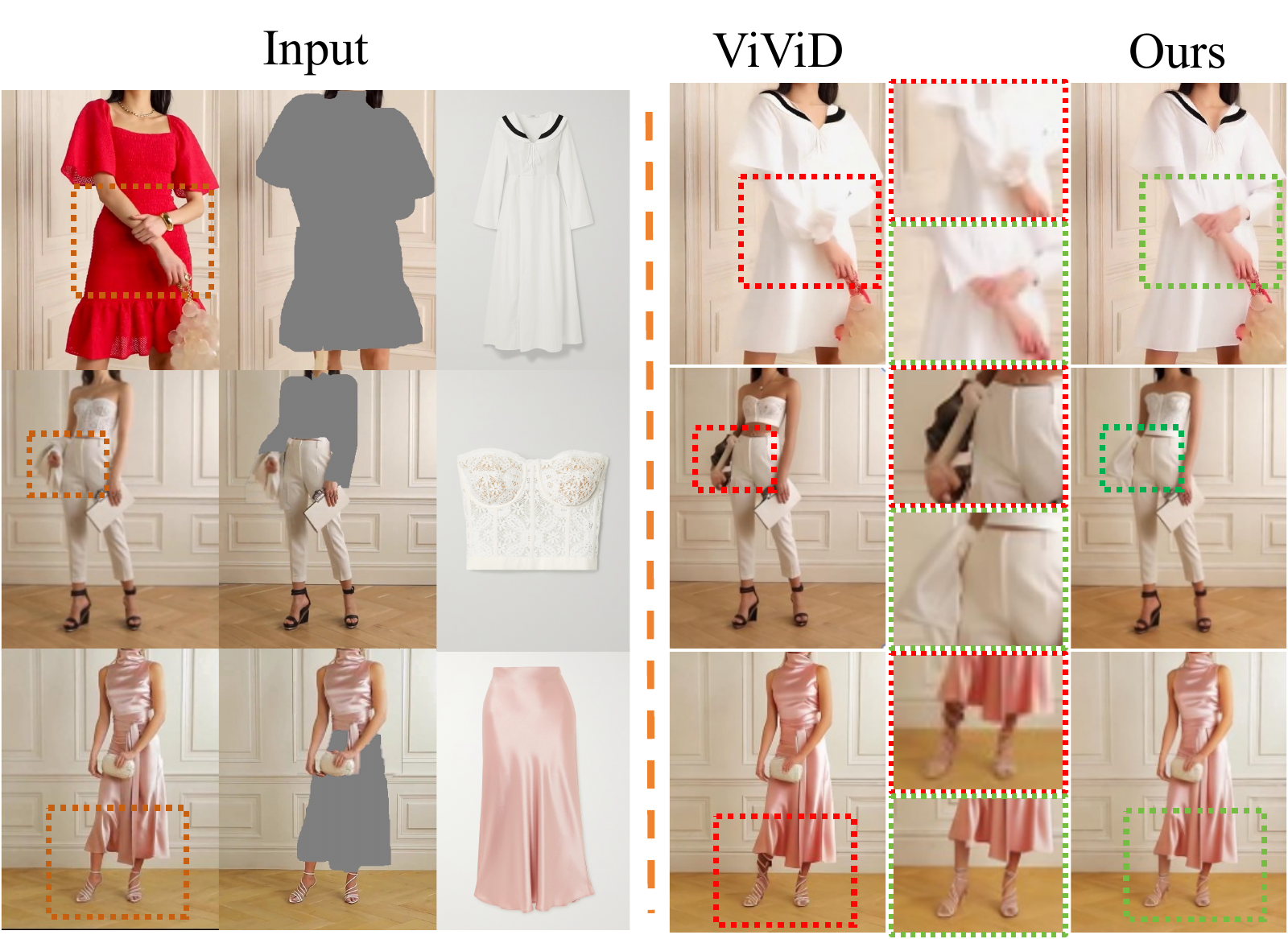}
    \caption{ChronoTailor effectively preserves information in non - edited regions.}
    \label{environment}
\end{minipage}
\vspace{-4mm}
\end{table}
\begin{figure}[t]
\begin{minipage}[c]{0.48\textwidth}
\begin{minipage}[t]{\textwidth}
    \centering
    \captionsetup{type=table}
    \caption{User study for the preference rate on the ViViD test dataset.}
    \label{table:user_study_vivid}
    \resizebox{\linewidth}{!}{ 
        \begin{tabular}{l p{1.2cm} p{1.8cm} p{1.8cm} p{1.8cm}}
            \toprule
            Method & Fidelity  &Background &Consistency & Overall\\
            \midrule
            OOTDiffusion \cite{xu2025ootdiffusion} & 15.36\% & 10.34\% & 0.00\% & 15.52\% \\
            StableVITON\cite{Kim_Gu_Park_Park_Choo_2023} & 2.51\% & 6.58\% & 0.00\% & 2.82\% \\
            CatVTON\cite{chong2024catvton} & 6.90\% & 9.09\% & 0.00\% & 6.58\% \\
            ViViD\cite{fang2024vivid} & 19.91\% & 21.79\% & 39.00\% & 20.22\% \\
            \textbf{ChronoTailor} & \textbf{55.33}\% & \textbf{52.19}\% & \textbf{61.00}\% & \textbf{54.86}\% \\
            \bottomrule
            \label{user_study}
        \end{tabular}
    }
\label{AMFE}
\end{minipage}
\begin{minipage}[t]{\textwidth}
\centering
    \captionsetup{type=table}
    \caption{Ablation study of Spatial-Temporal Attention Guidance}
    \label{ablation_stag}
    \resizebox{\textwidth}{!}{%
    \begin{tabular}{lccccc}
        \toprule
        Method & PSNR$\uparrow$ & SSIM$\uparrow$ & LPIPS$\downarrow$ & $VFID_{I3D}$$\downarrow$ & $VFID_{RES}$$\downarrow$ \\
        \midrule
        Baseline & 24.62 & 0.850 & 0.086 & 10.000 & 0.630 \\
        w/ RASG & 25.86 & 0.894 & 0.087 & 9.478 & 0.535 \\
        w/ ATFF & 26.52 & 0.899 & 0.084 & 9.834 & 0.612 \\
        w/ RASG + ATFF & \textbf{26.64} & \textbf{0.901} & \textbf{0.083} & \textbf{9.630} & \textbf{0.560} \\
        \bottomrule
    \end{tabular}
    }
\end{minipage}
\end{minipage}
\hfill
\begin{minipage}[c]{0.48\textwidth}
    \centering
    \includegraphics[width=\textwidth]{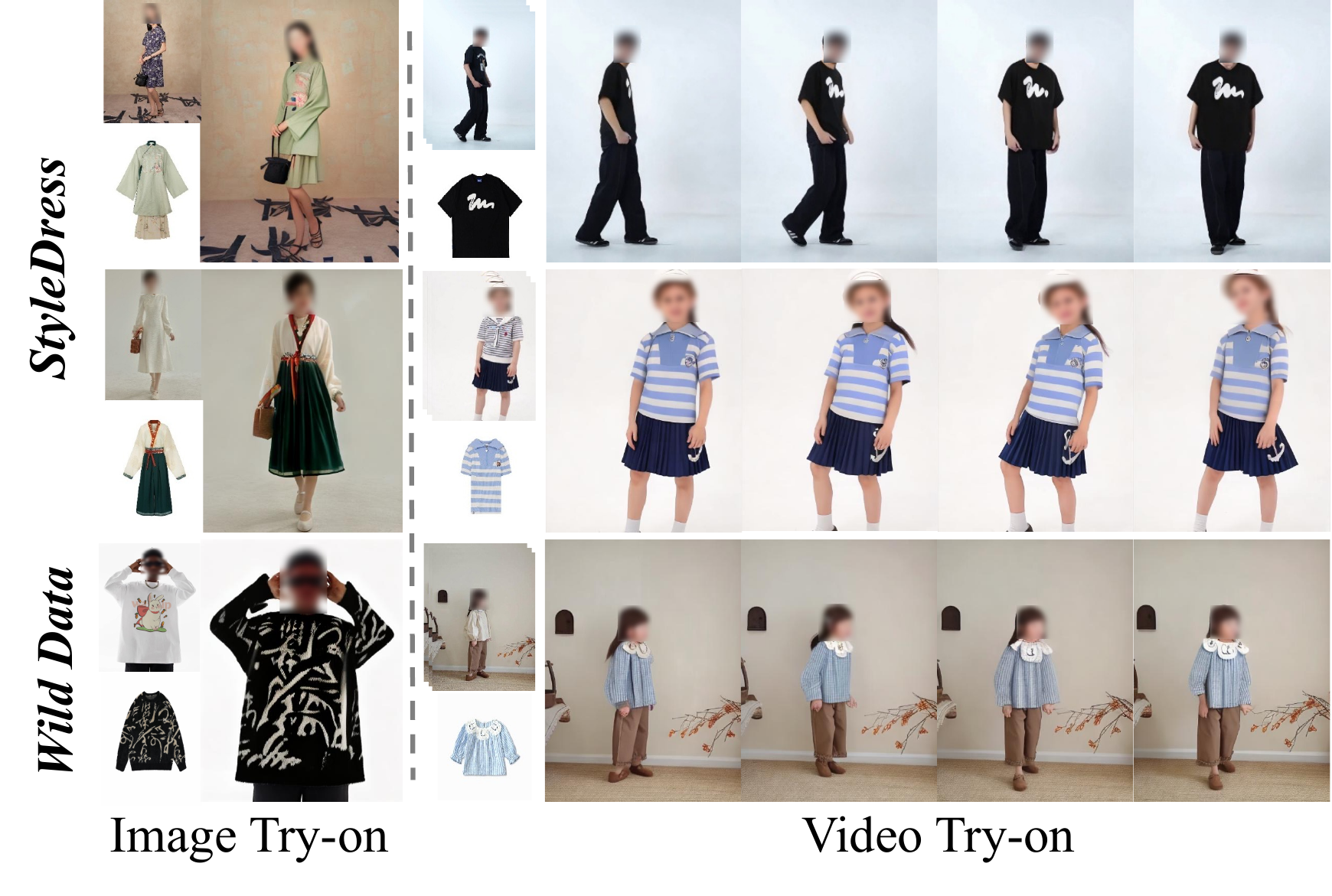}
    \caption{Qualitative results of our method on the StyleDress dataset.}
    \label{StyleDress}
\end{minipage}
\end{figure}

We evaluate our method on the video datasets VVT \cite{Dong_Liang_Shen_Wu_Chen_Yin_2019} and ViViD \cite{fang2024vivid}. 
For a fair comparison with baseline methods, we utilize models trained at a resolution of 512×384 pixels. To evaluate the virtual video try-on performance of our approach, we evaluate it in the VVT data set with a consistent resolution of 512×384 pixels.

\paragraph{Metrics and Baselines.} We adhere to the VTON video generation evaluation paradigm employing SSIM \cite{wang2004image}, LPIPS \cite{zhang2018unreasonable}, $VFID_{I3D}$ \cite{carreira2017quo} and $VFID_{Res}$ \cite{hara2018can} scores. Given the limited public availability of most comparative methods, our evaluation relies on reported results and accessible generated videos. We start our comparison with CP-VTON\cite{wang2018toward}, a GAN-based method, followed by comparisons with state-of-the-art diffusion-based methods: OOTDiffusion \cite{xu2025ootdiffusion}, Anydoor \cite{chen2024anydoor}, StableVITON\cite{Kim_Gu_Park_Park_Choo_2023} , CatVTON\cite{chong2024catvton}, and ViViD\cite{fang2024vivid}.
\paragraph{Implementation Details.} The experiments are conducted on two NVIDIA A800 GPUs. During the training process, all data is adjusted to a uniform resolution of 512×384. We set the batch size to 8 and the learning rate to $1e^{-5}$. More details are provided in Appendix~\ref{supp:trainingdetails}.
\vspace{-2mm}
\subsection{Qualitative Analysis}
Figure \ref{vvt} presents a visual comparison between our proposed method and baseline approaches using the VVT dataset. Further evidence of ChronoTailor's consistently superior quality and robustness in virtual try-on results is provided by the visualization results on the ViViD dataset, as shown in Figure \ref{visual_vivid}. Specifically, our method demonstrates enhanced preservation of garment texture details while maintaining inter-frame consistency, a significant challenge frequently encountered by other approaches that often struggle with the fidelity of clothing try-ons. In contrast, our method exhibits remarkable robustness, which can be attributed to our novel Spatial-Temporal Attention Guidance and Multi-scale Garment-Pose Feature Alignment mechanisms. In contrast to the ViViD method, our approach exhibits enhanced preservation of information beyond the edited regions, as depicted in Figure \ref{environment}. Furthermore, Figure \ref{StyleDress} showcases the effectiveness of ChronoTailor when evaluated on our collected StyleDress dataset and unseen in-the-wild data, thereby further validating its generalization. Appendix \ref{supp:full qualitative} presents additional visual comparison results between our method and other baselines. 
\subsection{Quantitative Comparisons}
\paragraph{Metric Evaluation.}Table \ref{VVT_DATA} presents a comparative analysis of our proposed method against baseline approaches on the VVT dataset. Our method demonstrates superior performance in terms of the SSIM and LPIPS scores, indicating improved visual fidelity. Furthermore, we achieve strong results in the $VFID_{Res}$ metric and exhibit competitive performance in $VFID_{I3D}$, collectively suggesting improved visual quality and temporal consistency. This robust performance effectively showcases the reliability of ChronoTailor for the task of virtual garment try-on in videos. Notably, while image-based methods can generate accurate single-frame results, they often suffer from flickering and inconsistencies due to limitations in temporal coherence and inter-frame motion handling.
\paragraph{User Study.}
We conduct a user study to evaluate the generation quality of our model. We utilize all test garments from the ViViD dataset and randomly present participants with 13 video virtual try-on results generated by both baseline methods and our approach. Each participant is asked to select their most preferred result based on four criteria: garment fidelity, background preservation, temporal consistency, and overall evaluation. We receive valid responses from 40 users, and the collected preferences are reported in Table \ref{user_study}. Across the four evaluation criteria, our proposed method is preferred by the majority of participants, achieving preference rates of $55.33\%$, $52.19\%$, $61.00\%$, and $54.86\%$, respectively. Furthermore, Appendix \ref{supp:User Study} presents further visual comparisons of the user study.
\subsection{Ablation Study}
\begin{figure}[t]
    \begin{minipage}[t]{0.48\textwidth}
        \centering
        \includegraphics[width=\textwidth]{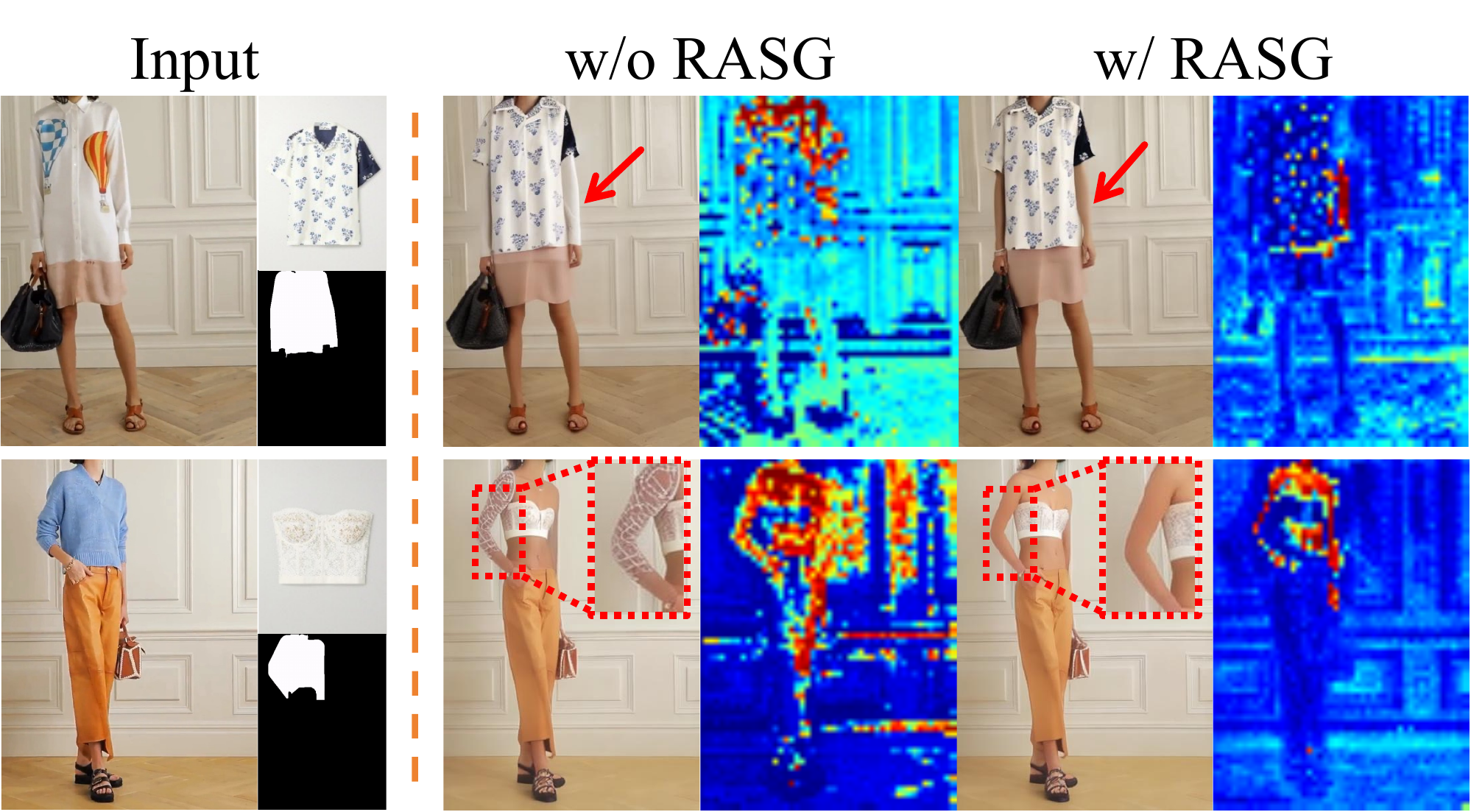}
        \caption{Ablation study of Region-Aware Spatial Guidance. Our method balances attention between editing and non - editing regions.}
        \label{RASG}
    \end{minipage}
    \hfill
    \begin{minipage}[t]{0.48\textwidth}
    \centering
    \includegraphics[width=\linewidth]{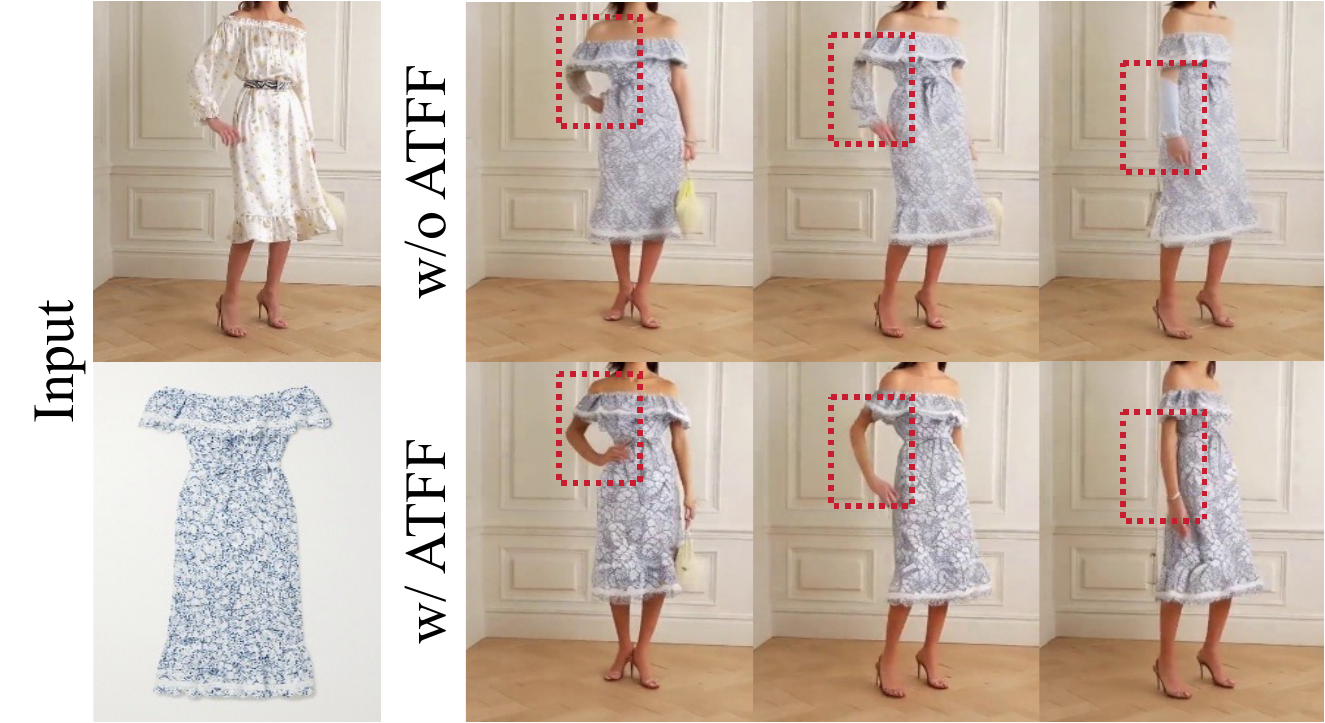}
    \caption{Ablation study of Attention-Driven Temporal Feature Fusion.}
    \label{ATFF}
    \end{minipage}
    \vspace{-3mm}
\end{figure}
\begin{figure}[t]
    \begin{minipage}[c]{0.49\textwidth}
        \centering
        \includegraphics[width=\textwidth]{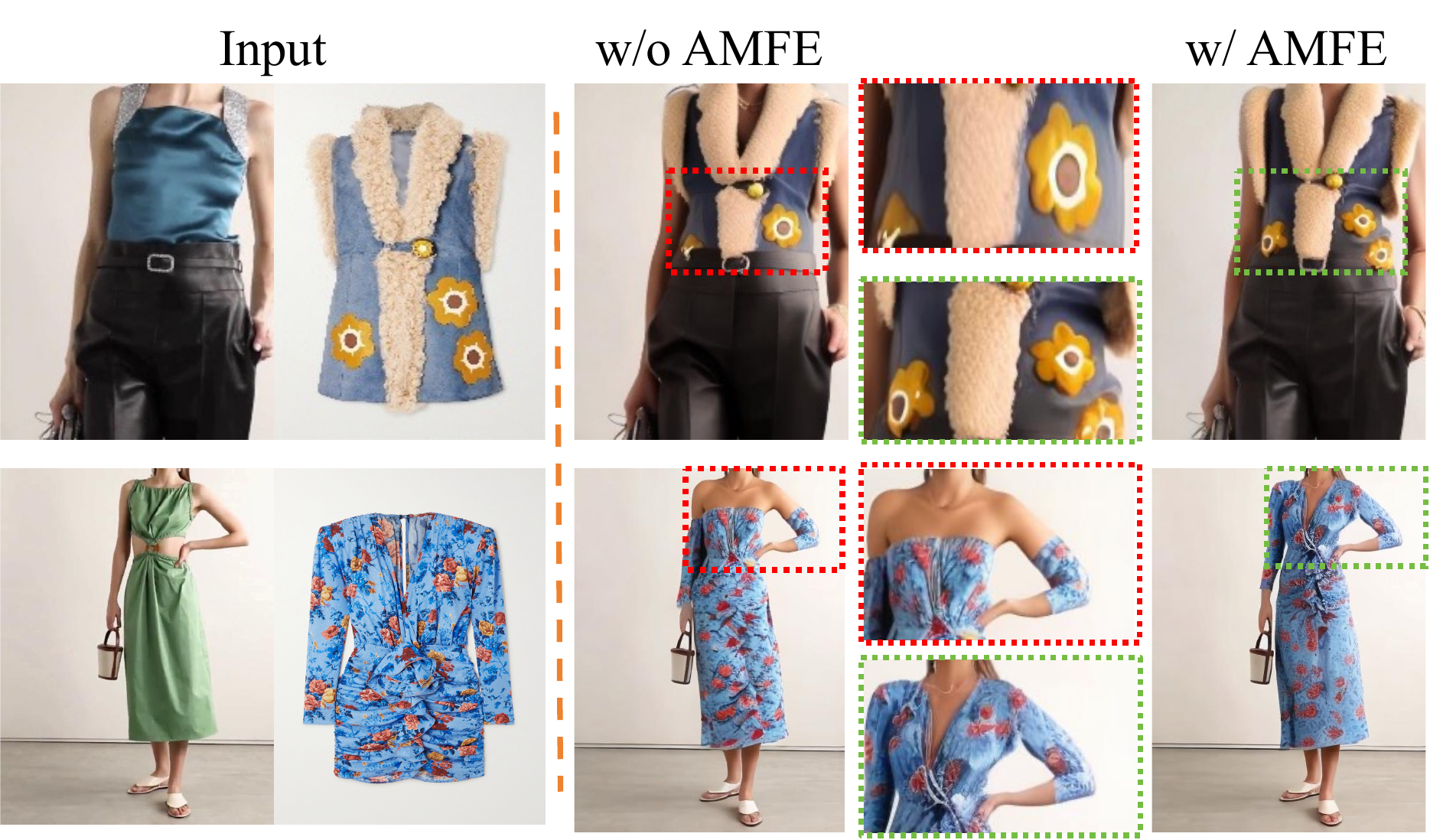}
        \caption{Ablation study of Adaptive Multi-scale Feature Extraction. The upper garment try-on fully recovers two floral patterns, demonstrating ChronoTailor’s capability to preserve intricate textures and semantic details.}
        \label{AMFE}
    \end{minipage}
    \hfill
    \begin{minipage}[c]{0.49\textwidth}
    \begin{minipage}[t]{\textwidth}
        \centering
        \captionsetup{type=table}
        \caption{Ablation study of Multi-scale Garment-Pose Feature Alignment.}
        \label{ablation_mgfa}
        \resizebox{\textwidth}{!}{%
        \begin{tabular}{lccccc}
            \toprule
            Method & PSNR$\uparrow$ & SSIM$\uparrow$ & LPIPS$\downarrow$ & $VFID_{I3D}$$\downarrow$ & $VFID_{RES}$$\downarrow$ \\
            \midrule
            Baseline & 24.62 & 0.850 & 0.086 & 10.000 & 0.630 \\
            w/ AMFE & 25.61 & 0.860 & 0.085 & 9.830 & 0.562 \\
            w/ GPFA & 26.23 & 0.904 & 0.085 & 9.855 & 0.574 \\
            w/ AMFE+GPFA & \textbf{26.26} & \textbf{0.904} & 0.084 & \textbf{9.739} & \textbf{0.553} \\
            \bottomrule
        \end{tabular}
        }
    \end{minipage}
    \begin{minipage}[t]{\textwidth}
        \centering
        \includegraphics[width=\textwidth]{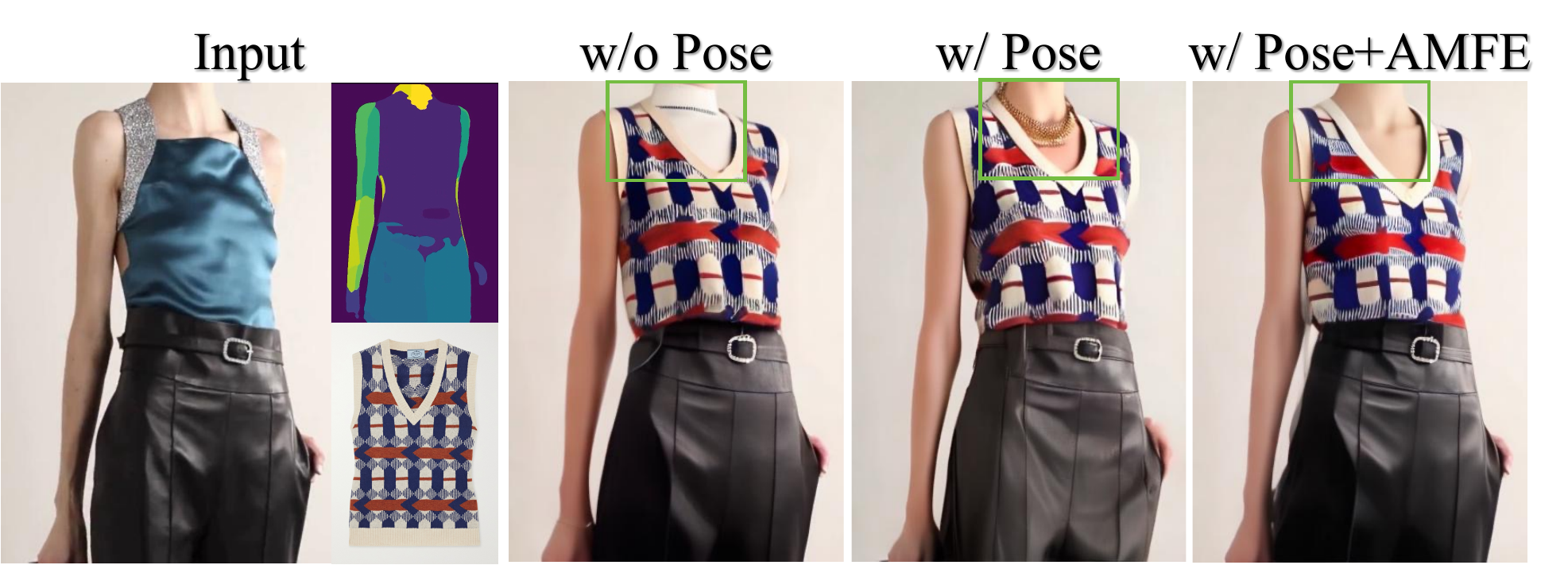} 
        \caption{Ablation study of Garment-Pose Feature Alignment. GPFA aligns garment features to poses, restoring occluded neck regions.}
        \label{GPFA}
    \end{minipage}
    \end{minipage}
    \vspace{-4mm}
\end{figure}
\paragraph{Effectiveness of RASG\&ATFF.}
To validate the efficacy of our spatial-temporal attention guidance, we first conduct an ablation study excluding the region-aware spatial guidance (RASG). As depicted in Figure \ref{RASG}, the integration of RASG demonstrably reduces the leakage of garment semantic information compared to the absence of RASG, indicating its effectiveness in constraining garment features to the appropriate regions. In Figure \ref{ATFF}, "w/o ATFF" indicates the absence of adaptive attention-driven temporal feature fusion (ATFF), leading to blurriness. Conversely, our ATFF recovers sharp textures by leveraging weighted temporal features to enhance the current frame's reconstruction, thereby improving temporal consistency and reducing video blur. The comparative results presented in Table \ref{ablation_stag} further substantiate the effectiveness of each proposed module. Appendix \ref{supp:Ablation Study_STAG} further presents a comprehensive validation of the overall effectiveness of the spatial-temporal attention guidance.
\paragraph{Effectiveness of AMFE\&GPFA.}
To validate our multi-scale garment-pose feature alignment, we performed ablation experiments using ReferenceNet-encoded garment features as the baseline.
Figure \ref{AMFE} shows that integrating the adaptive multi-scale feature extraction (AMFE) module significantly improves garment detail preservation and semantic reconstruction.
Figure \ref{GPFA} demonstrates the effectiveness of our garment-pose feature alignment. Incorporating Global-Local Pose Constraints (w/ Pose) enhances virtual try-on fidelity. Aligning AMFE-extracted multi-scale details with pose information (w/ Pose+AMFE) enables more precise pose-garment alignment, increasing try-on accuracy(see Appendix \ref{supp:Ablation Study_MG-PFA}). Quantitative results in Table \ref{ablation_mgfa} further confirm the substantial visual quality improvement of our method.
\section{Conclusion and Limitations}\label{sec:conclusion}
This paper presents ChronoTailor, a diffusion-based video virtual try-on framework enabling precise garment manipulation and fine-grained detail preservation via Spatial-Temporal Attention Guidance (segmentation masks, regularization, and cross-space attention for temporal coherence) and Multi-scale Garment-Pose Feature Alignment (hierarchical feature extraction and UNet-based pose alignment). We also introduce StyleDress, a high-quality dataset with diverse garments, environments, and poses, to be open-sourced post-publication. Extensive experiments show that ChronoTailor outperforms previous methods, generating temporally continuous videos while preserving garment details during motion.
\paragraph{Limitations.}Existing generative video virtual try-on methods heavily depend on high-fidelity segmentation and pose estimation, which degrade significantly under occlusion, low resolution, or high-dynamic motion. This often leads to attentional misalignment or spatial misregistration. While our method achieves precise garment alignment via spatial-temporal attention guidance, enhancing its generalization robustness in these complex scenarios remains a critical open challenge.

\medskip
{
\small
\bibliographystyle{plainnat} 
\bibliography{main} 

\begin{thebibliography}{54}
\providecommand{\natexlab}[1]{#1}
\providecommand{\url}[1]{\texttt{#1}}
\expandafter\ifx\csname urlstyle\endcsname\relax
  \providecommand{\doi}[1]{doi: #1}\else
  \providecommand{\doi}{doi: \begingroup \urlstyle{rm}\Url}\fi

\bibitem[Baldrati et~al.(2023)Baldrati, Morelli, Cartella, Cornia, Bertini, and Cucchiara]{baldrati2023multimodal}
Alberto Baldrati, Davide Morelli, Giuseppe Cartella, Marcella Cornia, Marco Bertini, and Rita Cucchiara.
\newblock Multimodal garment designer: Human-centric latent diffusion models for fashion image editing.
\newblock In \emph{Proceedings of the IEEE/CVF international conference on computer vision}, pages 23393--23402, 2023.

\bibitem[Carreira and Zisserman(2017)]{carreira2017quo}
Joao Carreira and Andrew Zisserman.
\newblock Quo vadis, action recognition? a new model and the kinetics dataset.
\newblock In \emph{Proceedings of the IEEE/CVF Conference on Computer Vision and Pattern Recognition}, pages 6299--6308, 2017.

\bibitem[Chen et~al.(2021)Chen, Lo, Huang, Shuai, and Cheng]{chen2021fashionmirror}
Chieh-Yun Chen, Ling Lo, Pin-Jui Huang, Hong-Han Shuai, and Wen-Huang Cheng.
\newblock Fashionmirror: Co-attention feature-remapping virtual try-on with sequential template poses.
\newblock In \emph{Proceedings of the IEEE/CVF International Conference on Computer Vision}, pages 13809--13818, 2021.

\bibitem[Chen et~al.(2024{\natexlab{a}})Chen, Huang, Liu, Shen, Zhao, and Zhao]{chen2024anydoor}
Xi~Chen, Lianghua Huang, Yu~Liu, Yujun Shen, Deli Zhao, and Hengshuang Zhao.
\newblock Anydoor: Zero-shot object-level image customization.
\newblock In \emph{Proceedings of the IEEE/CVF conference on computer vision and pattern recognition}, pages 6593--6602, 2024{\natexlab{a}}.

\bibitem[Chen et~al.(2024{\natexlab{b}})Chen, Liu, Chen, Feng, Liu, Shen, and Zhao]{chen2024livephoto}
Xi~Chen, Zhiheng Liu, Mengting Chen, Yutong Feng, Yu~Liu, Yujun Shen, and Hengshuang Zhao.
\newblock Livephoto: Real image animation with text-guided motion control.
\newblock In \emph{European Conference on Computer Vision}, pages 475--491. Springer, 2024{\natexlab{b}}.

\bibitem[Choi et~al.(2021)Choi, Park, Lee, and Choo]{Choi_Park_Lee_Choo_2021}
Seunghwan Choi, Sunghyun Park, Minsoo Lee, and Jaegul Choo.
\newblock Viton-hd: High-resolution virtual try-on via misalignment-aware normalization.
\newblock In \emph{Proceedings of IEEE/CVF Conference on Computer Vision and Pattern Recognition}, Jun 2021.

\bibitem[Choi et~al.(2024)Choi, Kwak, Lee, Choi, and Shin]{Choi_Kwak_Lee_Choi_Shin_2024}
Yisol Choi, Sangkyung Kwak, Kyungmin Lee, Hyungwon Choi, and Jinwoo Shin.
\newblock Improving diffusion models for virtual try-on.
\newblock Mar 2024.

\bibitem[Chong et~al.(2024)Chong, Dong, Li, Zhang, Zhang, Zhang, Zhao, Jiang, and Liang]{chong2024catvton}
Zheng Chong, Xiao Dong, Haoxiang Li, Shiyue Zhang, Wenqing Zhang, Xujie Zhang, Hanqing Zhao, Dongmei Jiang, and Xiaodan Liang.
\newblock Catvton: Concatenation is all you need for virtual try-on with diffusion models.
\newblock \emph{arXiv preprint arXiv:2407.15886}, 2024.

\bibitem[Cui et~al.(2021)Cui, McKee, and Lazebnik]{cui2021dressing}
Aiyu Cui, Daniel McKee, and Svetlana Lazebnik.
\newblock Dressing in order: Recurrent person image generation for pose transfer, virtual try-on and outfit editing.
\newblock In \emph{Proceedings of the IEEE/CVF international conference on computer vision}, pages 14638--14647, 2021.

\bibitem[Dong et~al.(2019)Dong, Liang, Shen, Wu, Chen, and Yin]{Dong_Liang_Shen_Wu_Chen_Yin_2019}
Haoye Dong, Xiaodan Liang, Xiaohui Shen, Bowen Wu, Bing-Cheng Chen, and Jian Yin.
\newblock Fw-gan: Flow-navigated warping gan for video virtual try-on.
\newblock In \emph{2019 IEEE/CVF International Conference on Computer Vision (ICCV)}, Oct 2019.

\bibitem[Esser et~al.(2023)Esser, Chiu, Atighehchian, Granskog, and Germanidis]{esser2023structure}
Patrick Esser, Johnathan Chiu, Parmida Atighehchian, Jonathan Granskog, and Anastasis Germanidis.
\newblock Structure and content-guided video synthesis with diffusion models.
\newblock In \emph{Proceedings of the IEEE/CVF international conference on computer vision}, pages 7346--7356, 2023.

\bibitem[Fang et~al.(2024)Fang, Zhai, Su, Song, Zhu, Wang, Chen, Liu, Cao, and Zha]{fang2024vivid}
Zixun Fang, Wei Zhai, Aimin Su, Hongliang Song, Kai Zhu, Mao Wang, Yu~Chen, Zhiheng Liu, Yang Cao, and Zheng-Jun Zha.
\newblock Vivid: Video virtual try-on using diffusion models.
\newblock \emph{arXiv preprint arXiv:2405.11794}, 2024.

\bibitem[Ge et~al.(2021)Ge, Song, Zhang, Ge, Liu, and Luo]{Ge_Song_Zhang_Ge_Liu_Luo_2021}
Yuying Ge, Yibing Song, Ruimao Zhang, Chongjian Ge, Wei Liu, and Ping Luo.
\newblock Parser-free virtual try-on via distilling appearance flows.
\newblock In \emph{Proceedings of IEEE/CVF Conference on Computer Vision and Pattern Recognition}, Jun 2021.

\bibitem[Goodfellow et~al.(2014)Goodfellow, Pouget-Abadie, Mirza, Xu, Warde-Farley, Ozair, Courville, and Bengio]{goodfellow2014generative}
Ian~J Goodfellow, Jean Pouget-Abadie, Mehdi Mirza, Bing Xu, David Warde-Farley, Sherjil Ozair, Aaron Courville, and Yoshua Bengio.
\newblock Generative adversarial nets.
\newblock \emph{Advances in neural information processing systems}, 27, 2014.

\bibitem[Gou et~al.(2023)Gou, Sun, Zhang, Si, Qian, and Zhang]{gou2023taming}
Junhong Gou, Siyu Sun, Jianfu Zhang, Jianlou Si, Chen Qian, and Liqing Zhang.
\newblock Taming the power of diffusion models for high-quality virtual try-on with appearance flow.
\newblock In \emph{Proceedings of the 31st ACM International Conference on Multimedia}, pages 7599--7607, 2023.

\bibitem[G{\"u}ler et~al.(2018)G{\"u}ler, Neverova, and Kokkinos]{guler2018densepose}
R{\i}za~Alp G{\"u}ler, Natalia Neverova, and Iasonas Kokkinos.
\newblock Densepose: Dense human pose estimation in the wild.
\newblock In \emph{Proceedings of the IEEE/CVF conference on computer vision and pattern recognition}, pages 7297--7306, 2018.

\bibitem[Guo et~al.(2023)Guo, Yang, Rao, Liang, Wang, Qiao, Agrawala, Lin, and Dai]{guo2023animatediff}
Yuwei Guo, Ceyuan Yang, Anyi Rao, Zhengyang Liang, Yaohui Wang, Yu~Qiao, Maneesh Agrawala, Dahua Lin, and Bo~Dai.
\newblock Animatediff: Animate your personalized text-to-image diffusion models without specific tuning.
\newblock \emph{arXiv preprint arXiv:2307.04725}, 2023.

\bibitem[Han et~al.(2018)Han, Wu, Wu, Yu, and Davis]{Han_Wu_Wu_Yu_Davis_2018}
Xintong Han, Zuxuan Wu, Zhe Wu, Ruichi Yu, and Larry~S. Davis.
\newblock Viton: An image-based virtual try-on network.
\newblock In \emph{Proceedings of IEEE/CVF Conference on Computer Vision and Pattern Recognition}, Jun 2018.

\bibitem[Hang et~al.(2023)Hang, Gu, Li, Bao, Chen, Hu, Geng, and Guo]{hang2023efficient}
Tiankai Hang, Shuyang Gu, Chen Li, Jianmin Bao, Dong Chen, Han Hu, Xin Geng, and Baining Guo.
\newblock Efficient diffusion training via min-snr weighting strategy.
\newblock In \emph{Proceedings of the IEEE/CVF international conference on computer vision}, pages 7441--7451, 2023.

\bibitem[Hara et~al.(2018)Hara, Kataoka, and Satoh]{hara2018can}
Kensho Hara, Hirokatsu Kataoka, and Yutaka Satoh.
\newblock Can spatiotemporal 3d cnns retrace the history of 2d cnns and imagenet?
\newblock In \emph{Proceedings of the IEEE/CVF conference on Computer Vision and Pattern Recognition}, pages 6546--6555, 2018.

\bibitem[Ho et~al.(2020)Ho, Jain, and Abbeel]{ho2020denoising}
Jonathan Ho, Ajay Jain, and Pieter Abbeel.
\newblock Denoising diffusion probabilistic models.
\newblock \emph{Advances in neural information processing systems}, 33:\penalty0 6840--6851, 2020.

\bibitem[Honda(2019)]{honda2019viton}
Shion Honda.
\newblock Viton-gan: Virtual try-on image generator trained with adversarial loss.
\newblock \emph{arXiv preprint arXiv:1911.07926}, 2019.

\bibitem[Hu(2024)]{hu2024animate}
Li~Hu.
\newblock Animate anyone: Consistent and controllable image-to-video synthesis for character animation.
\newblock In \emph{Proceedings of the IEEE/CVF Conference on Computer Vision and Pattern Recognition}, pages 8153--8163, 2024.

\bibitem[Hu et~al.(2022)Hu, Luo, and Chen]{hu2022make}
Yaosi Hu, Chong Luo, and Zhenzhong Chen.
\newblock Make it move: controllable image-to-video generation with text descriptions.
\newblock In \emph{Proceedings of the IEEE/CVF Conference on Computer Vision and Pattern Recognition}, pages 18219--18228, 2022.

\bibitem[Jiang et~al.(2024)Jiang, Hu, Luo, He, Xu, Peng, Zhang, Wang, Wu, and Fu]{jiang2024fitdit}
Boyuan Jiang, Xiaobin Hu, Donghao Luo, Qingdong He, Chengming Xu, Jinlong Peng, Jiangning Zhang, Chengjie Wang, Yunsheng Wu, and Yanwei Fu.
\newblock Fitdit: Advancing the authentic garment details for high-fidelity virtual try-on.
\newblock \emph{arXiv preprint arXiv:2411.10499}, 2024.

\bibitem[Jiang et~al.()Jiang, Wang, Yan, Liu, and Bigo]{Jiang_Wang_Yan_Liu_Bigo}
Jianbin Jiang, Tan Wang, He~Yan, Junhui Liu, and Bigo Bigo.
\newblock Clothformer: Taming video virtual try-on in all module.

\bibitem[Karras et~al.(2023)Karras, Holynski, Wang, and Kemelmacher-Shlizerman]{karras2023dreampose}
Johanna Karras, Aleksander Holynski, Ting-Chun Wang, and Ira Kemelmacher-Shlizerman.
\newblock Dreampose: Fashion image-to-video synthesis via stable diffusion.
\newblock In \emph{2023 IEEE/CVF International Conference on Computer Vision (ICCV)}, pages 22623--22633. IEEE, 2023.

\bibitem[Kim et~al.(2023)Kim, Gu, Park, Park, and Choo]{Kim_Gu_Park_Park_Choo_2023}
Jeongho Kim, Gyojung Gu, Minho Park, Sunghyun Park, and Jaegul Choo.
\newblock Stableviton: Learning semantic correspondence with latent diffusion model for virtual try-on.
\newblock Dec 2023.

\bibitem[Kuppa et~al.(2021)Kuppa, Jong, Liu, Liu, and Moh]{kuppa2021shineon}
Gaurav Kuppa, Andrew Jong, Xin Liu, Ziwei Liu, and Teng-Sheng Moh.
\newblock Shineon: Illuminating design choices for practical video-based virtual clothing try-on.
\newblock In \emph{Proceedings of the IEEE/CVF Winter Conference on Applications of Computer Vision}, pages 191--200, 2021.

\bibitem[Lee et~al.(2022)Lee, Gu, Park, Choi, and Choo]{lee2022high}
Sangyun Lee, Gyojung Gu, Sunghyun Park, Seunghwan Choi, and Jaegul Choo.
\newblock High-resolution virtual try-on with misalignment and occlusion-handled conditions.
\newblock In \emph{European Conference on Computer Vision}, pages 204--219. Springer, 2022.

\bibitem[Li et~al.(2023)Li, Li, Savarese, and Hoi]{li2023blip}
Junnan Li, Dongxu Li, Silvio Savarese, and Steven Hoi.
\newblock Blip-2: Bootstrapping language-image pre-training with frozen image encoders and large language models.
\newblock In \emph{International conference on machine learning}, pages 19730--19742. PMLR, 2023.

\bibitem[Li et~al.(2022)Li, Du, Wu, Xu, and Matusik]{li2022diffcloth}
Yifei Li, Tao Du, Kui Wu, Jie Xu, and Wojciech Matusik.
\newblock Diffcloth: Differentiable cloth simulation with dry frictional contact.
\newblock \emph{ACM Transactions on Graphics (TOG)}, 42\penalty0 (1):\penalty0 1--20, 2022.

\bibitem[Liu et~al.(2025)Liu, Sun, Di, Sun, Yang, Zou, and Bao]{liu2025moee}
Huaize Liu, Wenzhang Sun, Donglin Di, Shibo Sun, Jiahui Yang, Changqing Zou, and Hujun Bao.
\newblock Moee: Mixture of emotion experts for audio-driven portrait animation.
\newblock \emph{arXiv preprint arXiv:2501.01808}, 2025.

\bibitem[Morelli et~al.(2022)Morelli, Fincato, Cornia, Landi, Cesari, and Cucchiara]{morelli2022dress}
Davide Morelli, Matteo Fincato, Marcella Cornia, Federico Landi, Fabio Cesari, and Rita Cucchiara.
\newblock Dress code: High-resolution multi-category virtual try-on.
\newblock In \emph{Proceedings of the IEEE/CVF conference on computer vision and pattern recognition}, pages 2231--2235, 2022.

\bibitem[Morelli et~al.(2023)Morelli, Baldrati, Cartella, Cornia, Bertini, and Cucchiara]{morelli2023ladi}
Davide Morelli, Alberto Baldrati, Giuseppe Cartella, Marcella Cornia, Marco Bertini, and Rita Cucchiara.
\newblock Ladi-vton: Latent diffusion textual-inversion enhanced virtual try-on.
\newblock In \emph{Proceedings of the 31st ACM international conference on multimedia}, pages 8580--8589, 2023.

\bibitem[Nguyen et~al.(2024)Nguyen, Nguyen, Nguyen, and Nguyen]{nguyen2024swifttry}
Hung Nguyen, Quang Qui-Vinh Nguyen, Khoi Nguyen, and Rang Nguyen.
\newblock Swifttry: Fast and consistent video virtual try-on with diffusion models.
\newblock \emph{arXiv preprint arXiv:2412.10178}, 2024.

\bibitem[Ravi et~al.(2024)Ravi, Gabeur, Hu, Hu, Ryali, Ma, Khedr, R{\"a}dle, Rolland, Gustafson, et~al.]{ravi2024sam}
Nikhila Ravi, Valentin Gabeur, Yuan-Ting Hu, Ronghang Hu, Chaitanya Ryali, Tengyu Ma, Haitham Khedr, Roman R{\"a}dle, Chloe Rolland, Laura Gustafson, et~al.
\newblock Sam 2: Segment anything in images and videos.
\newblock \emph{arXiv preprint arXiv:2408.00714}, 2024.

\bibitem[Rombach et~al.(2022)Rombach, Blattmann, Lorenz, Esser, and Ommer]{rombach2022high}
Robin Rombach, Andreas Blattmann, Dominik Lorenz, Patrick Esser, and Bj{\"o}rn Ommer.
\newblock High-resolution image synthesis with latent diffusion models.
\newblock In \emph{Proceedings of the IEEE/CVF conference on computer vision and pattern recognition}, pages 10684--10695, 2022.

\bibitem[Ronneberger et~al.(2015)Ronneberger, Fischer, and Brox]{ronneberger2015u}
Olaf Ronneberger, Philipp Fischer, and Thomas Brox.
\newblock U-net: Convolutional networks for biomedical image segmentation.
\newblock In \emph{Medical image computing and computer-assisted intervention--MICCAI 2015: 18th international conference, Munich, Germany, October 5-9, 2015, proceedings, part III 18}, pages 234--241. Springer, 2015.

\bibitem[Sun et~al.(2024)Sun, Li, Di, Liang, Zhang, Li, Chen, and Cui]{sun2024uniavatar}
Wenzhang Sun, Xiang Li, Donglin Di, Zhuding Liang, Qiyuan Zhang, Hao Li, Wei Chen, and Jianxun Cui.
\newblock Uniavatar: Taming lifelike audio-driven talking head generation with comprehensive motion and lighting control.
\newblock \emph{arXiv preprint arXiv:2412.19860}, 2024.

\bibitem[Wang et~al.(2018)Wang, Zheng, Liang, Chen, Lin, and Yang]{wang2018toward}
Bochao Wang, Huabin Zheng, Xiaodan Liang, Yimin Chen, Liang Lin, and Meng Yang.
\newblock Toward characteristic-preserving image-based virtual try-on network.
\newblock In \emph{Proceedings of the European conference on computer vision (ECCV)}, pages 589--604, 2018.

\bibitem[Wang et~al.(2004)Wang, Bovik, Sheikh, and Simoncelli]{wang2004image}
Zhou Wang, Alan~C Bovik, Hamid~R Sheikh, and Eero~P Simoncelli.
\newblock Image quality assessment: from error visibility to structural similarity.
\newblock \emph{IEEE transactions on image processing}, 13\penalty0 (4):\penalty0 600--612, 2004.

\bibitem[Wei et~al.(2025)Wei, Yu, Zhou, and Wang]{wei20253dv}
Min Wei, Chaohui Yu, Jingkai Zhou, and Fan Wang.
\newblock 3dv-ton: Textured 3d-guided consistent video try-on via diffusion models.
\newblock \emph{arXiv preprint arXiv:2504.17414}, 2025.

\bibitem[Xiong et~al.(2024)Xiong, Wu, Tan, Li, Tang, Chen, Li, Ma, and Li]{xiong2024sam2}
Xinyu Xiong, Zihuang Wu, Shuangyi Tan, Wenxue Li, Feilong Tang, Ying Chen, Siying Li, Jie Ma, and Guanbin Li.
\newblock Sam2-unet: Segment anything 2 makes strong encoder for natural and medical image segmentation.
\newblock \emph{arXiv preprint arXiv:2408.08870}, 2024.

\bibitem[Xu et~al.(2025)Xu, Gu, Chen, and Chen]{xu2025ootdiffusion}
Yuhao Xu, Tao Gu, Weifeng Chen, and Arlene Chen.
\newblock Ootdiffusion: Outfitting fusion based latent diffusion for controllable virtual try-on.
\newblock In \emph{Proceedings of the AAAI Conference on Artificial Intelligence}, volume~39, pages 8996--9004, 2025.

\bibitem[Xu et~al.(2024)Xu, Chen, Wang, Xing, Zhai, Sang, Lan, Xiao, and Gao]{xu2024tunnel}
Zhengze Xu, Mengting Chen, Zhao Wang, Linyu Xing, Zhonghua Zhai, Nong Sang, Jinsong Lan, Shuai Xiao, and Changxin Gao.
\newblock Tunnel try-on: Excavating spatial-temporal tunnels for high-quality virtual try-on in videos.
\newblock In \emph{Proceedings of the 32nd ACM International Conference on Multimedia}, pages 3199--3208, 2024.

\bibitem[Xu et~al.()Xu, Zhang, Liew, Yan, Liu, Zhang, Feng, and Shou]{Xu_Zhang_Liew_Yan_Liu_Zhang_Feng_Shou}
Zhongcong Xu, Jianfeng Zhang, JunHao Liew, Hanshu Yan, Jia-Wei Liu, Chenxu Zhang, Jiashi Feng, and MikeZheng Shou.
\newblock Magicanimate: Temporally consistent human image animation using diffusion model.

\bibitem[Zeng et~al.(2024)Zeng, Song, Nie, Tian, Wang, and Liu]{zeng2024cat}
Jianhao Zeng, Dan Song, Weizhi Nie, Hongshuo Tian, Tongtong Wang, and An-An Liu.
\newblock Cat-dm: Controllable accelerated virtual try-on with diffusion model.
\newblock In \emph{Proceedings of the IEEE/CVF conference on computer vision and pattern recognition}, pages 8372--8382, 2024.

\bibitem[Zhang et~al.(2025)Zhang, Wu, Sun, Liu, Di, Chen, and Zou]{zhang2025semanticlatentmotionportrait}
Qiyuan Zhang, Chenyu Wu, Wenzhang Sun, Huaize Liu, Donglin Di, Wei Chen, and Changqing Zou.
\newblock Semantic latent motion for portrait video generation, 2025.

\bibitem[Zhang et~al.(2018)Zhang, Isola, Efros, Shechtman, and Wang]{zhang2018unreasonable}
Richard Zhang, Phillip Isola, Alexei~A Efros, Eli Shechtman, and Oliver Wang.
\newblock The unreasonable effectiveness of deep features as a perceptual metric.
\newblock In \emph{Proceedings of the IEEE/CVF conference on computer vision and pattern recognition}, pages 586--595, 2018.

\bibitem[Zhang(2018)]{zhang2018improved}
Zijun Zhang.
\newblock Improved adam optimizer for deep neural networks.
\newblock In \emph{2018 IEEE/ACM 26th international symposium on quality of service (IWQoS)}, pages 1--2. Ieee, 2018.

\bibitem[Zhong et~al.(2021)Zhong, Wu, Tan, Lin, and Wu]{Zhong_Wu_Tan_Lin_Wu_2021}
Xiaojing Zhong, Zhonghua Wu, Taizhe Tan, Guosheng Lin, and Qingyao Wu.
\newblock Mv-ton: Memory-based video virtual try-on network.
\newblock In \emph{Proceedings of the 29th ACM International Conference on Multimedia}, Oct 2021.

\bibitem[Zhu et~al.(2023)Zhu, Yang, Zhu, Reda, Chan, Saharia, Norouzi, and Kemelmacher-Shlizerman]{zhu2023tryondiffusion}
Luyang Zhu, Dawei Yang, Tyler Zhu, Fitsum Reda, William Chan, Chitwan Saharia, Mohammad Norouzi, and Ira Kemelmacher-Shlizerman.
\newblock Tryondiffusion: A tale of two unets.
\newblock In \emph{Proceedings of the IEEE/CVF Conference on Computer Vision and Pattern Recognition}, pages 4606--4615, 2023.

\bibitem[Zhu et~al.(2024)Zhu, Chen, Dai, Dong, Xu, Cao, Yao, Zhu, and Zhu]{zhu2024champ}
Shenhao Zhu, Junming~Leo Chen, Zuozhuo Dai, Zilong Dong, Yinghui Xu, Xun Cao, Yao Yao, Hao Zhu, and Siyu Zhu.
\newblock Champ: Controllable and consistent human image animation with 3d parametric guidance.
\newblock In \emph{European Conference on Computer Vision}, pages 145--162. Springer, 2024.

\end{thebibliography}
}

\newpage

\appendix

\section{APPENDIX}\label{supp:implement details}





\subsection{Preliminary}\label{supp:Preliminary} 
The Diffusion Model represents a category of generative models that synthesize data via an iterative denoising process from a foundation of random noise. In order to enhance the proficiency of training and inference, Stable Diffusion (SD) \cite{rombach2022high} functions within the compact latent space of a pre-trained autoencoder, denoted as $\bar{D}(\bar{\mathcal{E}}(\cdot))$, where $\bar{\mathcal{E}}$ and $\bar{\mathcal{D}}$ symbolize the encoder and decoder, respectively. Considering an input image $I_{0}$ along with its corresponding latent feature $x_{0} =\bar{\mathcal{E}}(I_{0})$, the diffusion forward process is delineated as follows:
\begin{equation}
\mathbf{x}_t = \sqrt{\bar{\alpha}_t} \mathbf{x}_0 + \sqrt{1 - \bar{\alpha}_t} \mathbf{\epsilon}, \quad \mathbf{\epsilon} \sim \mathcal{N}(\mathbf{0}, \mathbf{I}),
\end{equation}
here, \(\bar{\alpha}_t = \prod_{t_{0}=1}^{t}(1 - \bar{\beta}_{t_{0}})\) and \(\bar{\beta}_{t_{0}}\) represents the pre-specified variance schedule at timestep \({t_{0}}\). 

During the diffusion backward process, referred to as $L_{LDM}$ and conditioned on the textual context $\bar{c}$, the model is engineered to anticipate the noise $\epsilon_{\theta}(\mathbf{x}_t, \bar{c}, t)$, which can be summarized as:
\begin{equation}
\mathcal{L}_{\text{LDM}} = \mathbb{E}_{x_0,\mathbf{\epsilon},\bar{c},t}\|\mathbf{\epsilon} - \mathbf{\epsilon}_{\theta}(x_t,\bar{c}, t)\|^2 .
\end{equation}

Where $\mathbf{\epsilon}_{\theta}$ stands for a U-Net model that is trained to anticipate the noise within the diffusion process.
\subsection{Training and Inference Details}\label{supp:trainingdetails}
The backbone network initializes the UNet and clothing encoder with Stable Diffusion-1.5 weights, while keeping the VAE encoder and CLIP image encoder weights unchanged. The temporal module uses weights from the AnimeDiff \cite{guo2023animatediff} motion module for initialization. Our model is optimized using the Adam optimizer~\cite{zhang2018improved} with \(\beta_1 = 0.9\), \(\beta_2 = 0.999\), and a fixed learning rate applied consistently across image pretraining and video fine-tuning stages. We integrate the Min-SNR Weighting Strategy~\cite{hang2023efficient} with \(\gamma = 5.0\) to enhance training stability, and employ mixed-precision FP16 throughout training to optimize computational efficiency.

The training pipeline consists of two phases: an initial image pretraining of 60,000 steps on image datasets (240 GPU hours) focused on spatial feature learning, followed by video fine-tuning for an additional 30,000 steps on video data (120 GPU hours), where only the temporal module is trained to optimize temporal coherence while preserving spatial accuracy, utilizing a continuous 24-frame sequence sampled for video training.

For inference, tests are conducted on an NVIDIA GeForce RTX 4090. The model processes 48 frames per video at \(512 \times 384\) resolution to balance efficiency and fidelity.
\subsection{Video Generation Model Details}

\subsubsection{Region-Aware Spatial Guidance}
Notably, RASG is exclusively applied during the training phase and introduces no additional computational overhead at inference. This design ensures that our approach maintains high efficiency for real-time applications while significantly enhancing the visual realism and semantic consistency of virtual try-on results during training. By decoupling guidance mechanism usage from inference, RASG achieves a practical balance between training-time optimization and deployment efficiency.

\section{Additional Analysis for ChronoTailor}
\subsection{Additional Qualitative Results}\label{supp:full qualitative}
\subsubsection{Qualitative results on ViViD and VVT Datasets}\label{supp:VIVID and VVT}
\begin{figure}[htbp] 
    \centering
    \includegraphics[width=\textwidth]{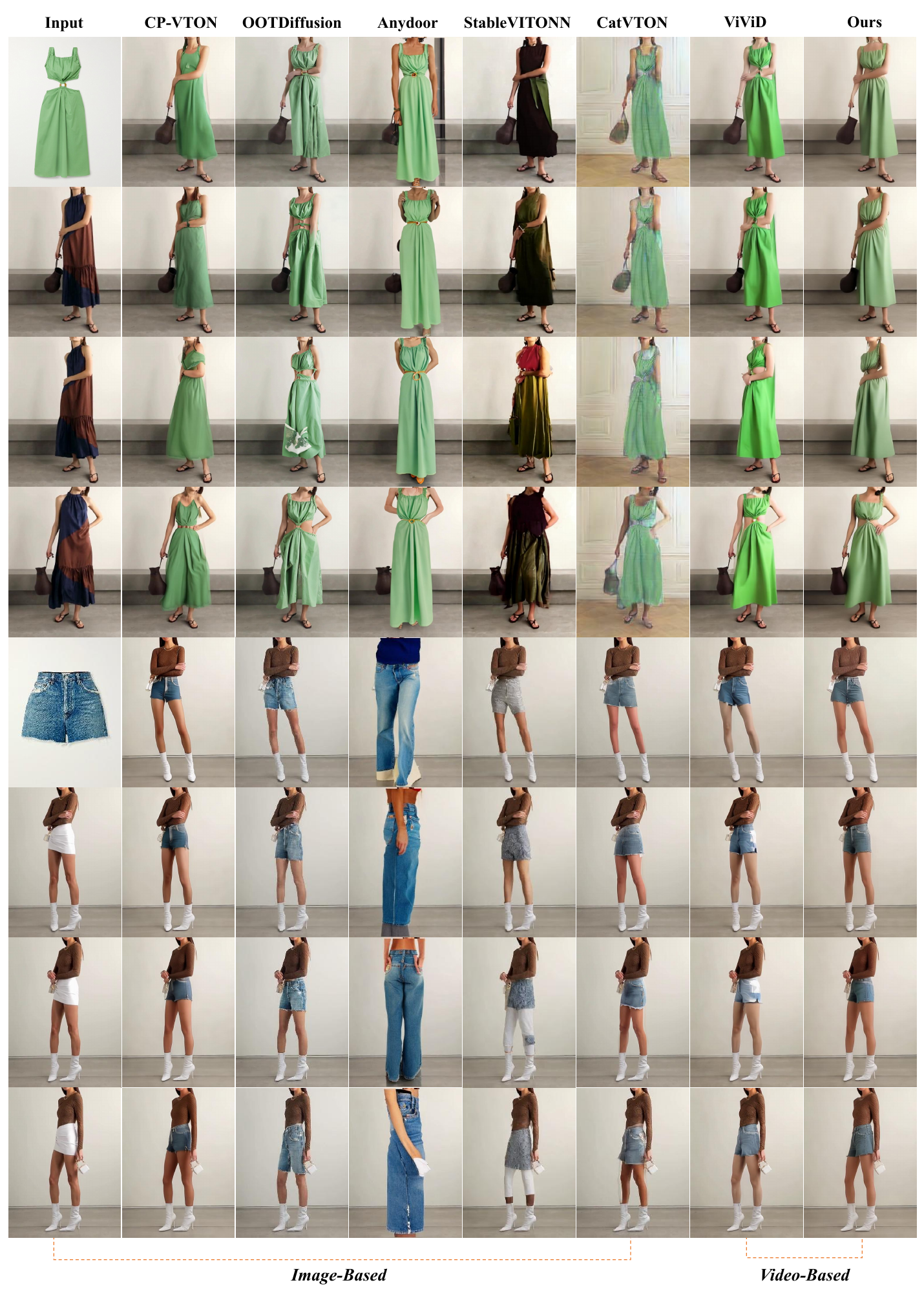}
    \caption{Qualitative comparison on the ViViD dataset. Our method can generate more color-accurate fitting results.}
    \label{vivid_supp}
\end{figure}

\begin{figure}[htbp]  
    \centering
    \includegraphics[width=\textwidth]{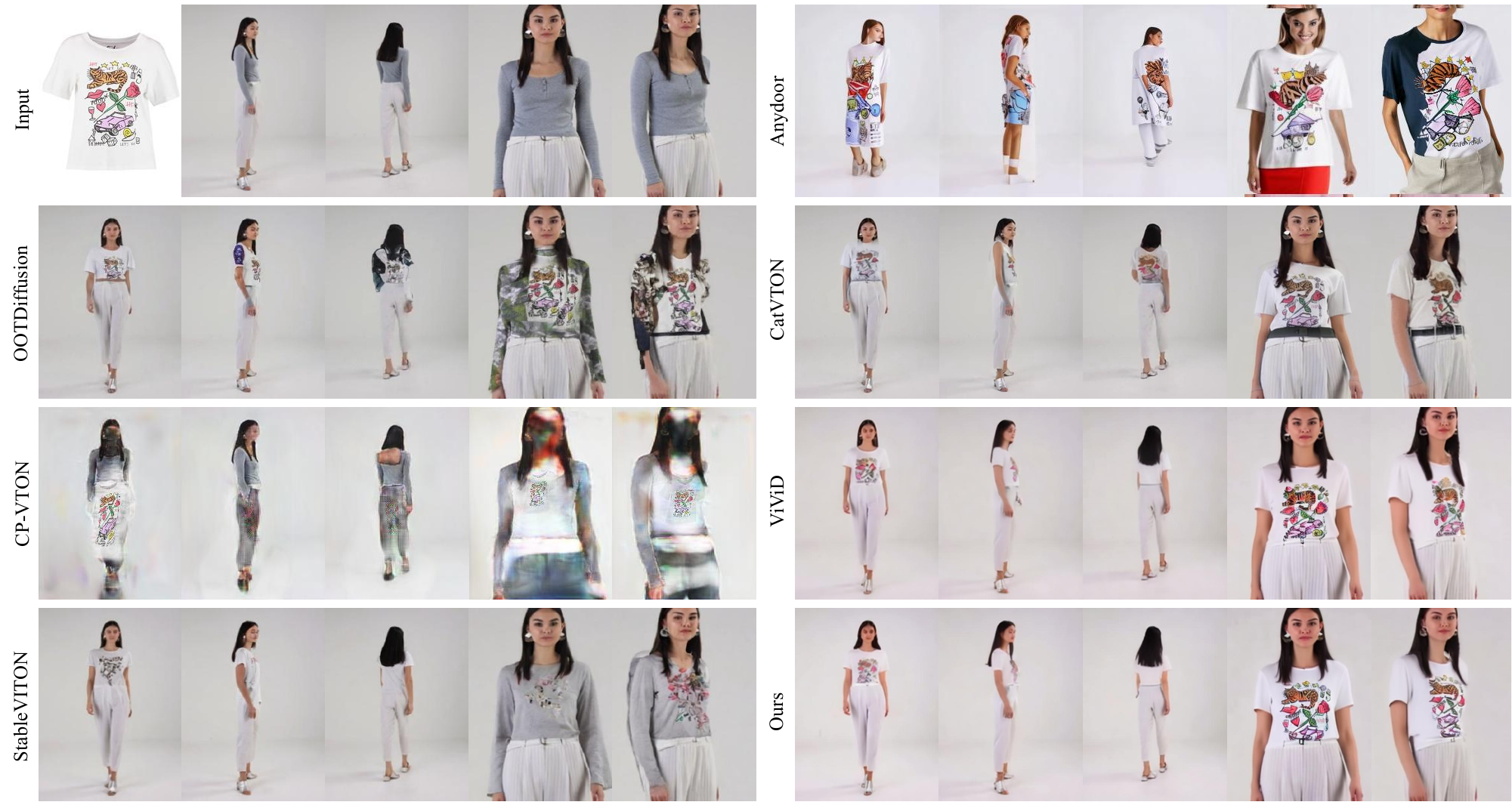}
    \caption{Qualitative comparison on the VVT dataset. Our method generates clearer texture details in the editing region.}
    \label{vvt_supp}
\end{figure}

\paragraph{Qualitative results on the ViViD dataset.}
In Figure \ref{vivid_supp}, we present the video virtual try-on results of models on the VIVID dataset \cite{fang2024vivid}. The GAN-based model \cite{wang2018toward}exhibits color discrepancies in skin tones and clothing textures. OOTDiffusion \cite{xu2025ootdiffusion} incorrectly generates ultra-shorts into mid-length shorts, while AnyDoor \cite{chen2024anydoor} struggles to achieve precise garment replacement. Both StableVITON \cite{Kim_Gu_Park_Park_Choo_2023} and CatVTON \cite{chong2024catvton} produce clothing category errors—for example, transforming shorts into long pants or skirts. ViViD not only shows color deviations but also introduces spurious patterns absent from the input.
\paragraph{Qualitative results on the VVT dataset.}
The experimental results on the VVT dataset\cite{Dong_Liang_Shen_Wu_Chen_Yin_2019} demonstrate our model's capability to restore texture details while maintaining excellent pose alignment, as shown in Figure \ref{vvt_supp}.

\subsubsection{Qualitative results on the StyleDress dataset}\label{supp:StyleDress}
\begin{figure*}[htbp]  
    \centering
    \includegraphics[width=\textwidth]{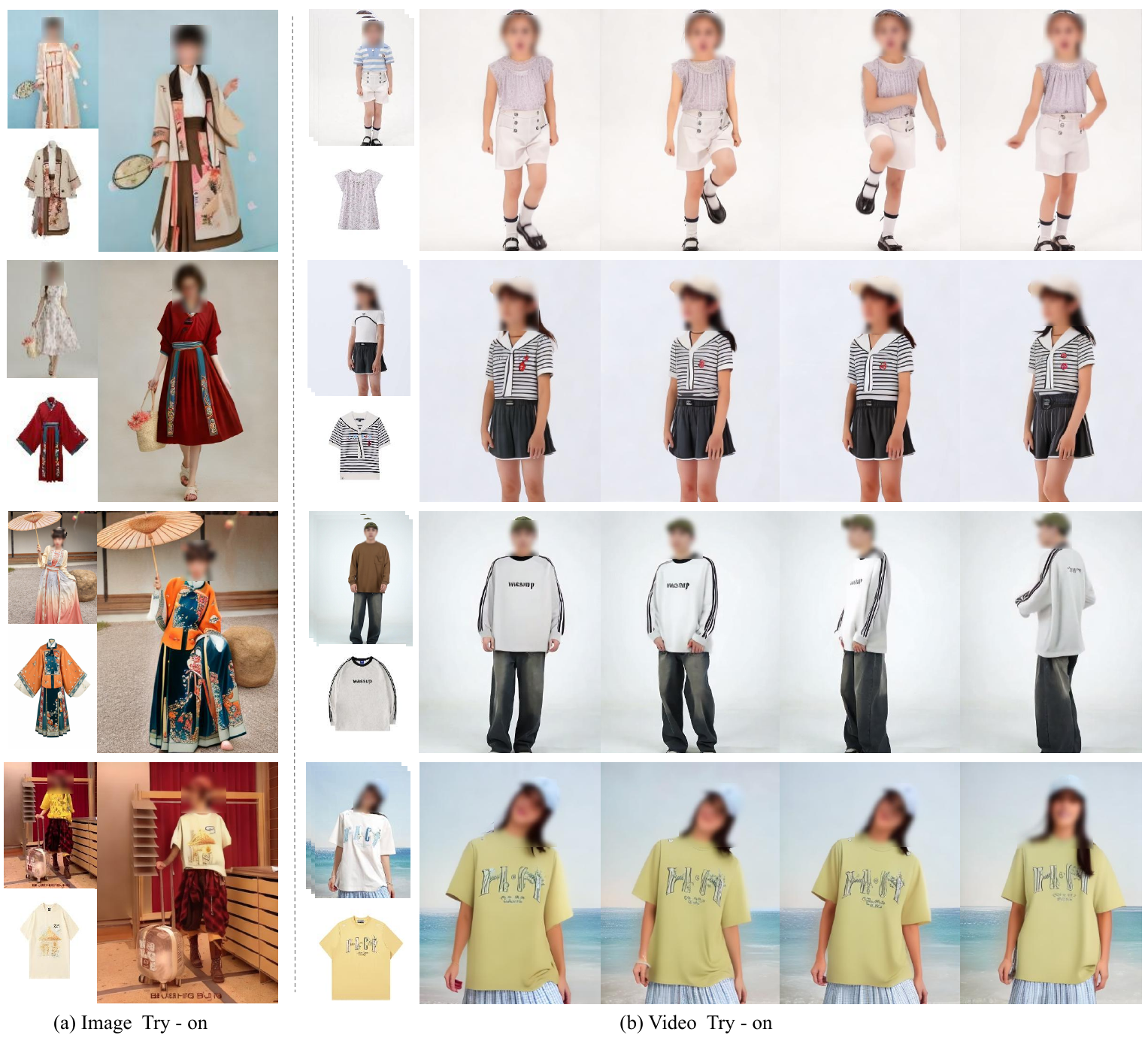} 
    \caption{Additional video try-on results on the StyleDress dataset are shown. Left column image try-ons for complex textures demonstrate precise pattern mapping via dataset-enhanced textures. Right column video try-ons show: 1) semantic alignment for small body types (first two rows); 2) spatiotemporal texture consistency during walking/turning motions (third row); 3) background integrity preservation during clothing replacement (fourth row).}
    \label{styledress_supp_result}
\end{figure*}

Figure \ref{styledress_supp_result} showcases additional results of our video virtual try-on method on the StyleDress dataset. By leveraging the dataset’s multimodal data (images and videos), our approach demonstrates superior performance compared to methods trained on the VVT dataset. Specifically, our method ensures consistent garment texture quality across frames, enabled by StyleDress' rich information on clothing types, backgrounds, and poses, highlighting the dataset's unique value for robust, high-fidelity virtual try-on systems.

\subsubsection{Qualitative results on the VITON-HD Dataset}\label{supp:VITON-HD}
\begin{figure*}[htbp] 
    \centering
    \includegraphics[width=\textwidth]{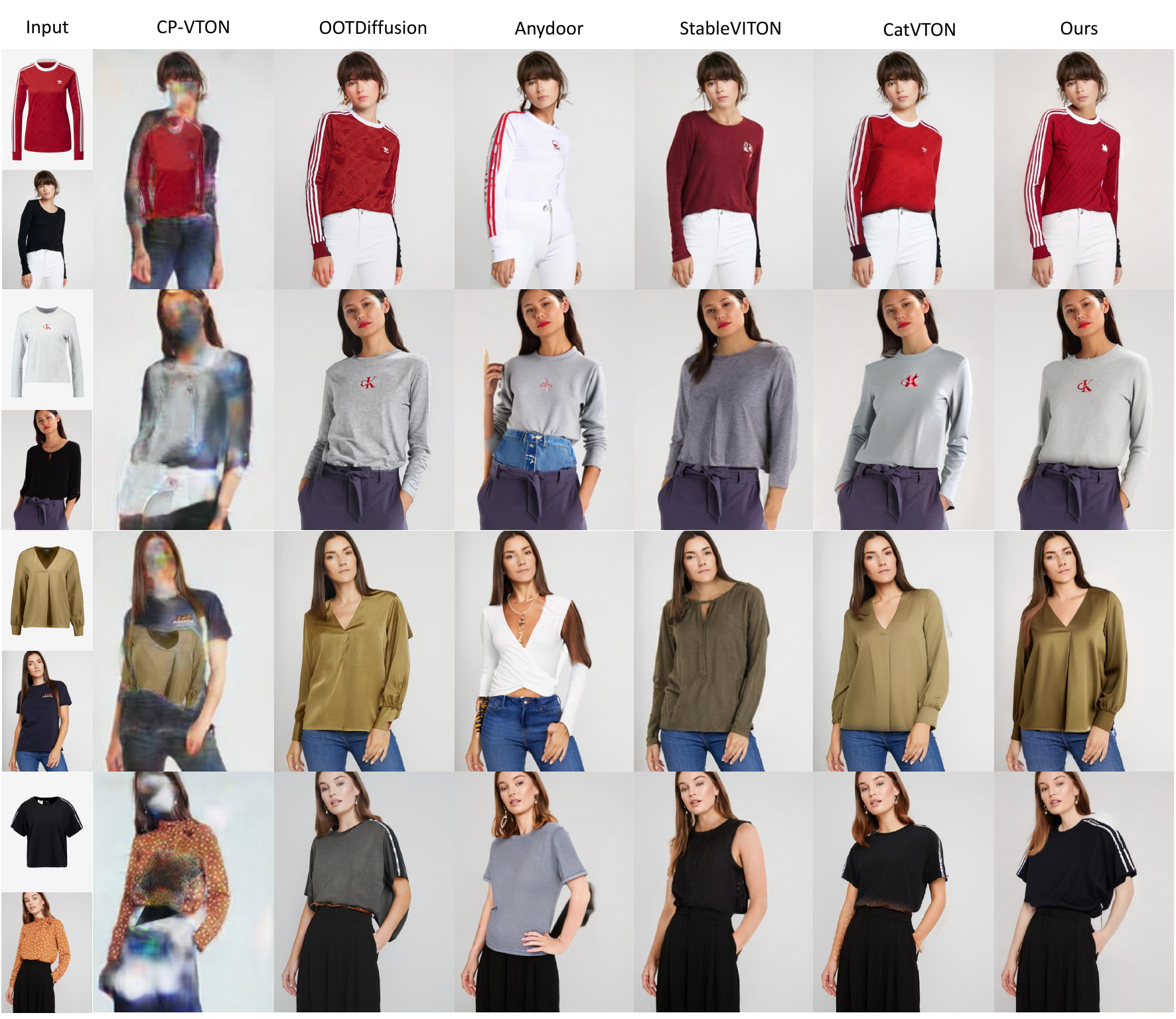} 
    \caption{Qualitative comparison on the VITON-HD dataset.}
    \label{viton_supp}
\end{figure*}

Our approach is also effective for image virtual try-on tasks. As shown in Figure \ref{viton_supp}, the results on the VITON-HD \cite{lee2022high} dataset demonstrate that our method achieves competitive performance in image-based garment transfer, balancing visual realism and clothing detail preservation.


\subsection{StyleDress}\label{supp:StyleDress}
Our \textbf{StyleDress multimodal dataset} addresses key limitations in existing virtual try-on datasets, particularly concerning gender bias, limited pose variations, and restricted body shape representation. We developed this high-resolution, video-centric dataset, augmented with an image dataset, to provide comprehensive training materials for realistic virtual try-on in diverse, real-world scenarios.

The video dataset features multi-action indoor scenes (frame rate \(\ge\) 24fps) capturing complex dynamics like walking and dancing, alongside a wide array of clothing categories. This design specifically enhances the model's ability to learn spatio-temporal consistency. Complementing this, the image dataset, compiled from e-commerce platforms and fashion social networks, boasts diverser material categories and design elements, emphasizing high-resolution texture details for increased data diversity.
StyleDress significantly expands diversity by including a full gender spectrum, varied body shapes (BMI 16-28), a wide range of human postures, and complex indoor and outdoor backgrounds. This broad coverage of real-world complexities, such as dynamic lighting, multi-angle camera views, and articulated human motions, highlights the dataset's robust advantages over traditional collections.

\begin{figure*}[htbp] 
    \centering
    \includegraphics[width=\textwidth]{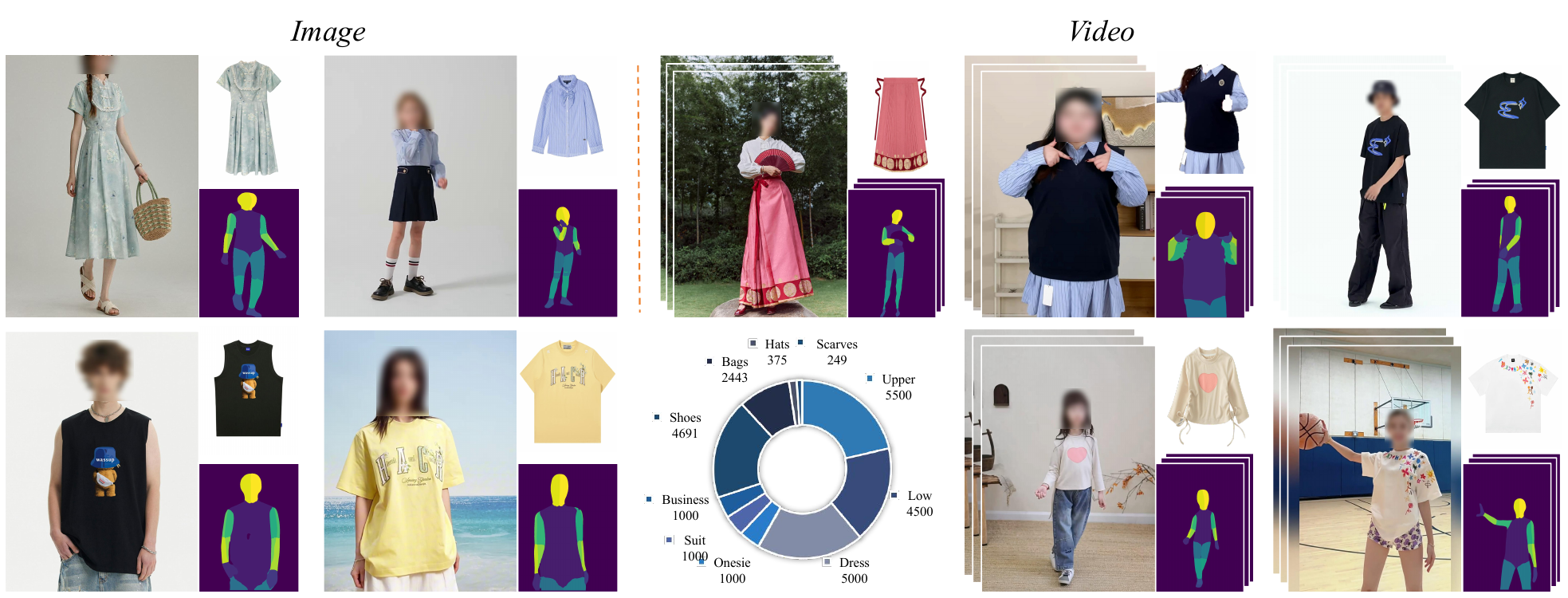} 
    \caption{Our StyleDress dataset offers additional examples with diverse clothing, challenging poses and environments, and balanced representation of body shapes, genders, and ages.}
    \label{styledress_V2}
\end{figure*}

The dataset employs a five-tuple annotation system and a rigorous processing pipeline. Videos are preprocessed and cropped to \(832 \times 624\) pixels, utilizing DensePose \cite{guler2018densepose} for 3D pose and surface semantics and OOTDiffusion for agnostic masks. Clothing images are precisely segmented using SAM2 \cite{ravi2024sam}, followed by manual refinement after BLIP-2 \cite{li2023blip} classification, which also added 7 new subcategories like jumpsuits. To further enhance semantic retention in non-edited areas, over 5000 accessory samples from Net-A-Porter are included with annotated retention masks in occluded regions. A hierarchical clothing attribute label system, constructed through BLIP-2 and manual annotation, further enriches the dataset. Figure \ref{styledress_V2} showcases typical samples from this dataset, which will be open-sourced for public research.

\subsection{User Study Details}\label{supp:User Study}
\begin{figure*}[htbp] 
    \centering
    \includegraphics[width=\textwidth]{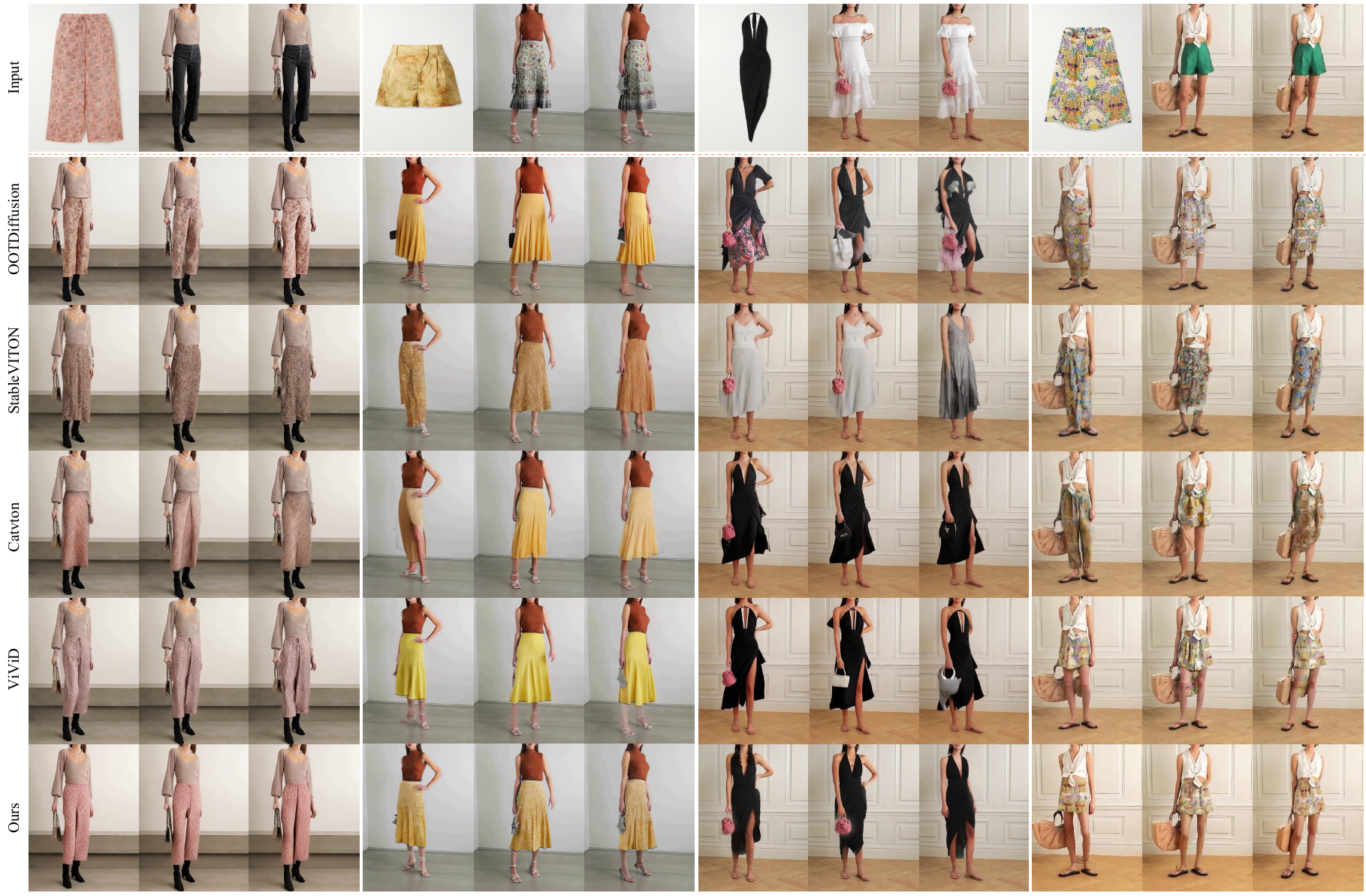} 
    \caption{ChronoTailor best preserves the texture details and appearance information of garments across different poses.}
    \label{userstudy}
\end{figure*}

In this section, we conduct a comprehensive user study through qualitative evaluation to compare ChronoTailor with existing methods (OOTDiffusion \cite{xu2025ootdiffusion}, Anydoor \cite{chen2024anydoor}, StableVITON \cite{Kim_Gu_Park_Park_Choo_2023}, CatVTON \cite{chong2024catvton}, and ViViD \cite{fang2024vivid}), as illustrated in Figure \ref{userstudy}.
Our user study utilizes 13 reference garment images from the ViViD test set and their corresponding virtual try-on video results, with exemplars visualized in Figure 10. A total of 50 evaluation samples are collected. All methods are anonymized as Method \textcircled{1}, \textcircled{2} , \textcircled{3} , \textcircled{4} and \textcircled{5}, and the presentation order of try-on video results is randomized in each comparative group to eliminate order bias.
The evaluation dimensions and their operational definitions are as follows:
\begin{itemize}
    \item \textbf{Fidelity}: The faithfulness of garment appearance reproduction in try-on results.
    \item \textbf{Background}: The retention of environmental details, accessories, and unedited regions (e.g., hands).
    \item \textbf{Consistency}: The smoothness and coherence of video sequences across frames.
    \item \textbf{Overall}: A comprehensive assessment of videos' holistic generation quality, integrating visual fidelity, temporal coherence, and background preservation to evaluate realism and technical integrity.
\end{itemize}
This randomized and anonymized evaluation framework ensures objective and unbiased qualitative comparisons, enabling rigorous assessment of ChronoTailor’s performance against state-of-the-art alternatives in terms of visual fidelity, background preservation (of unedited regions), temporal consistency, and overall visual quality.

\subsection{More Ablation Study}\label{supp:More Ablation Study}
\subsubsection{Effectiveness of Spatial-Temporal Attention Guidance}\label{supp:Ablation Study_STAG}
\begin{figure*}[htbp]
    \centering
    \includegraphics[width=0.5\linewidth]{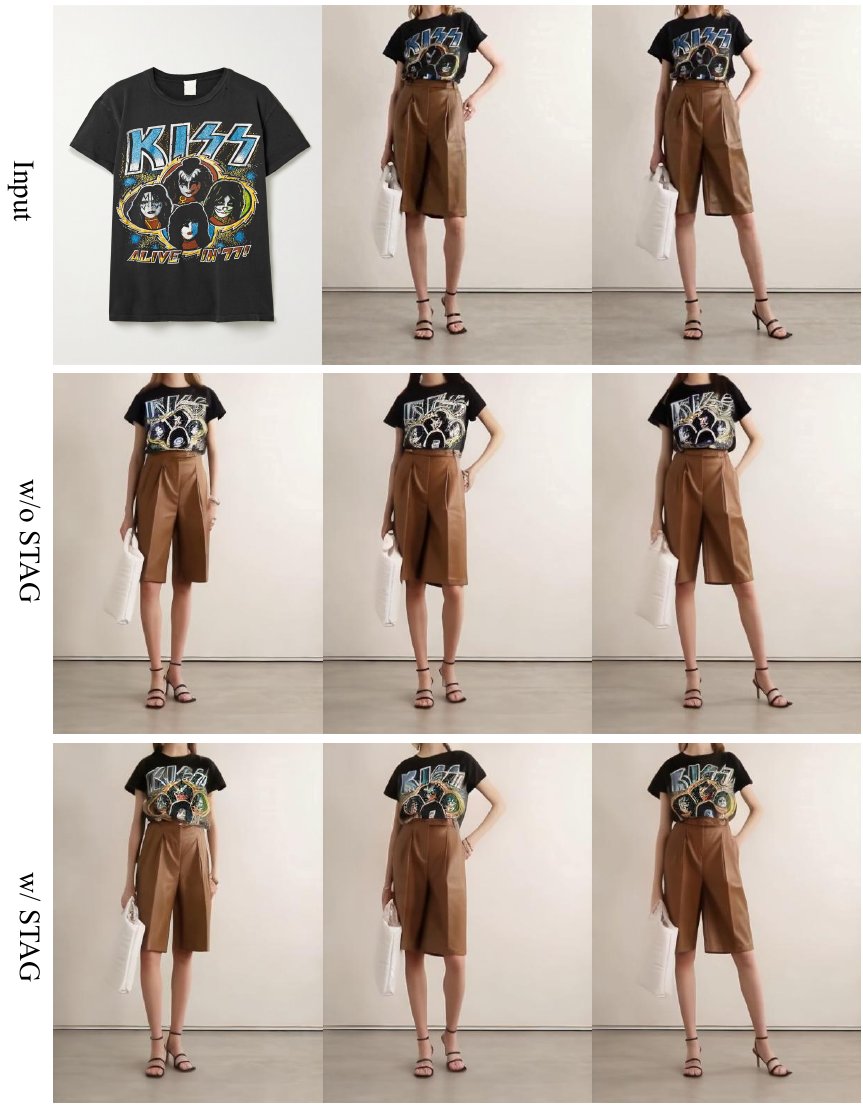}
    \caption{Ablation study of Spatial-Temporal Attention Guidance. Our method not only achieves superior semantic information preservation but also further reduces color discrepancy.}
    \label{STAG_supp}
\end{figure*}

In Figure \ref{STAG_supp}, we present more visual results to validate the effectiveness of Spatial-Temporal Attention Guidance (STAG). "w/o STAG" denotes the case without adopting Spatial-Temporal Attention Guidance, while "w/ STAG" denotes the case with adopting it. "w/o STAG" encounters severe color difference issues, resulting in significant loss of semantic information in clothing colors. In contrast, "w/ STAG" remarkably reduces color differences and improves clothing consistency. This is attributed to the fact that our designed Spatial-Temporal Attention Guidance adjusts the attention regions from both spatial and temporal dimensions, thereby providing more precise guidance for the feature integration process.

\subsubsection{Effectiveness of Multi-scale Garment-Pose Feature Alignment}\label{supp:Ablation Study_MG-PFA}
\begin{figure*}[htbp] 
    \centering
    \includegraphics[width=0.5\textwidth]{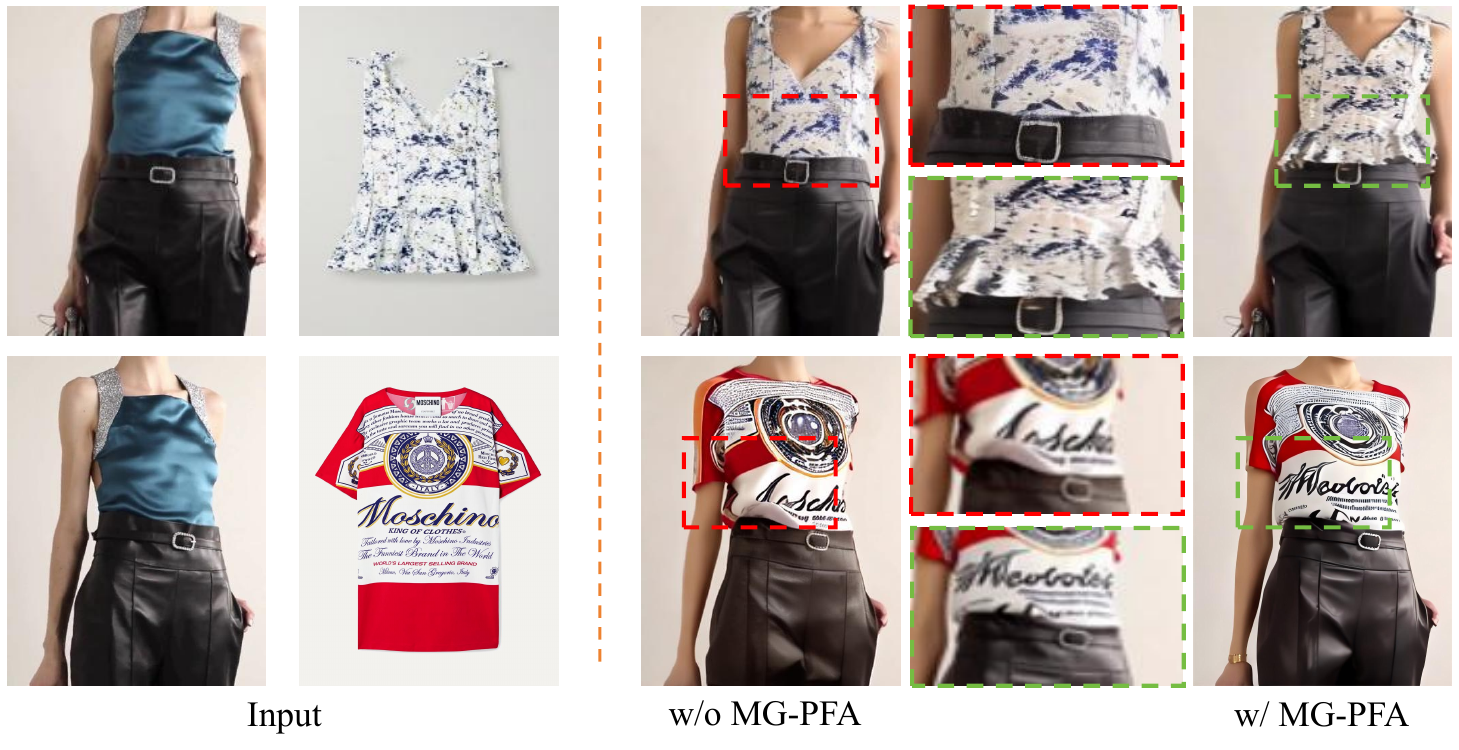} 
    \caption{Ablation study of Multi-scale Garment-Pose Feature Alignment. ChronoTailor achieves precise retention of garment details through deep semantic modeling, significantly maintaining the hierarchical texture of upper garment lace and the visual integrity of text.}
    \label{MG-PFA_supp}
\end{figure*}

Figure \ref{MG-PFA_supp} shows additional visual results validate the effectiveness of the Multi-Scale Garment Position Alignment (MG-PFA). The baseline model without MG-PFA (w/o MG-PFA, Base model) exhibits significant geometric distortion and semantic information loss in garments, manifested as contour warping and texture fragmentation. In contrast, the model with MG-PFA (w/ MG-PFA) achieves precise alignment between garment semantics and human poses through cross-layer interaction of multi-scale features, significantly enhancing geometric consistency and detail fidelity. Experimental results demonstrate that MG-PFA mitigates deformation defects caused by traditional single-scale alignment via multi-resolution hierarchical alignment, effectively suppressing garment distortion under complex poses.

\newpage

\end{document}